\pdfoutput=1
\documentclass[11pt]{article}

\usepackage[final]{acl}

\usepackage{amsmath}
\usepackage{multirow}
\usepackage{bm}
\usepackage{makecell}
\usepackage{wrapfig}
\usepackage{float}
\usepackage[ruled,linesnumbered]{algorithm2e}
\usepackage{algpseudocode}
\usepackage{pifont}

\usepackage{siunitx}
\usepackage{colortbl}

\usepackage{caption}
\usepackage[utf8]{inputenc} % allow utf-8 input
\usepackage[T1]{fontenc}    % use 8-bit T1 fonts
\usepackage{hyperref}       % hyperlinks
\usepackage{url}            % simple URL typesetting
\usepackage{booktabs}       % professional-quality tables
\usepackage{amsfonts}       % blackboard math symbols
\usepackage{graphicx}       % include graphics
\usepackage{nicefrac}       % compact symbols for 1/2, etc.
\usepackage{microtype}      % microtypography
\usepackage{xcolor}         % colors

\usepackage{enumitem}

\usepackage{times}
\usepackage{latexsym}
\usepackage{subfigure}
\usepackage{pdfpages}
\usepackage{multicol}
\usepackage{graphicx}
\usepackage{bm}
\usepackage{color}
\usepackage{multirow}
\definecolor{brown}{RGB}{139,64,0}
\definecolor{pink}{RGB}{255,170,182}
\definecolor{purple}{RGB}{160,32,240}
\AtBeginDocument{%
  }

\usepackage[T1]{fontenc}
\usepackage[utf8]{inputenc}

\usepackage{microtype}

\usepackage{inconsolata}
\usepackage{graphicx}

\title{GLProtein: Global-and-Local Structure Aware Protein Representation Learning}

\author{
 \textbf{Yunqing Liu\textsuperscript{\textdagger}}~~~~ 
 \textbf{Wenqi Fan\thanks{Corresponding author}\textsuperscript{\textdagger~\textdaggerdbl}}~~~~  
 \textbf{Xiao-Yong Wei\textsuperscript{\textdagger~\S}}~~~~  
 \textbf{Qing Li\textsuperscript{\textdagger}} 
\\
 \textsuperscript{\textdagger}Department of Computing (COMP), The Hong Kong Polytechnic University\\ 
 \textsuperscript{\textdaggerdbl}Department of Management and Marketing (MM), The Hong Kong Polytechnic University\\ 
 \textsuperscript{\S}Department of Computer Science, Sichuan University
\\
\small{
 \texttt{
   \href{mailto:yunqing617.liu@connect.polyu.hk}{yunqing617.liu@connect.polyu.hk}
 }}
 \\
\small{
 \texttt{wenqifan03@gmail.com, 
   \{cs007.wei, csqli\}@comp.polyu.edu.hk
 }
}
}

\begin{document}

\maketitle

\begin{abstract}

Proteins are central to biological systems, participating as building blocks across all forms of life. Despite advancements in understanding protein functions through protein sequence analysis, there remains potential for further exploration in integrating protein structural information. We argue that the structural information of proteins is not only limited to their 3D information but also encompasses information from amino acid molecules (local information) to protein-protein structure similarity (global information). To address this, we propose \textbf{GLProtein}, the first framework in protein pre-training that incorporates both global structural similarity and local amino acid details to enhance prediction accuracy and functional insights. GLProtein innovatively combines protein-masked modelling with triplet structure similarity scoring, protein 3D distance encoding and substructure-based amino acid molecule encoding. Experimental results demonstrate that GLProtein outperforms previous methods in several bioinformatics tasks, including predicting protein-protein interactions, contact prediction, and so on. 
% The code is available at \url{https://anonymous.4open.science/r/GLProtein-9F2C/}.

\end{abstract}

\section{Introduction}

Proteins are fundamental to virtually every biological process, serving as the building blocks for cells and organs and acting as catalysts, messengers, and structural elements in all life forms. 
Understanding the structure and function of proteins is crucial for advances in health, agriculture, and environmental science, making protein research a cornerstone of biotechnology and medicinal science~\cite{ding2019selective,davis2024food,zhao2024holistic}.
Recognizing the critical role of proteins in various scientific fields, many efforts have been made to design computational methods to further understand these crucial molecules~\cite{weng2021late,zhao2020exploring}.
% \wq{***This transition is not smooth enough!}
Particularly, protein representation learning, as one significant part, involves capturing the complex features and relationships within proteins in a condensed form that can be utilized for various computational tasks and analyses.
It is crucial for enhancing the understanding of protein structures and functions, improving predictive modelling in bioinformatics, facilitating the drug discovery process, and advancing our knowledge of biological systems through interpretable and efficient representations of proteins~\cite{somnath2021multi,liu2023generative,gao2024drugclip}. 

\begin{figure}[t]
    \centering
    \includegraphics[width=1\linewidth]{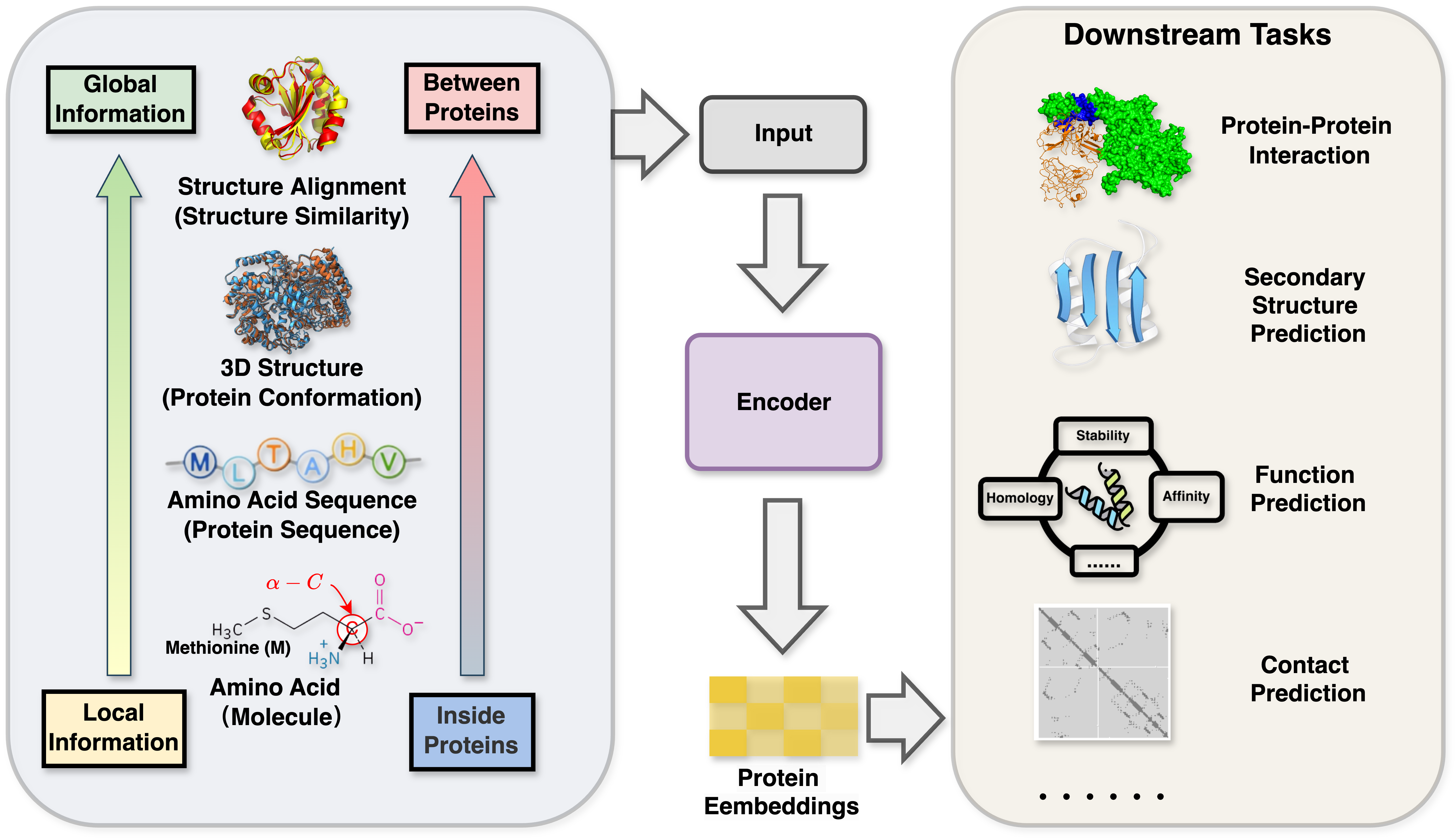}
    \vspace{-5mm}
    \caption{An illustration on protein representation learning flow. Protein information from local information (inside proteins) to global information (between proteins) can be used as input. 
    This input undergoes encoding by a protein encoder to generate a protein representation across various downstream tasks.}
    \label{fig:flow}
    \vskip -0.2in
\end{figure}

In recent years, the success of language models in natural language processing (NLP) has paved the way for innovative approaches in bioinformatics areas, such as protein modeling~\cite{xiao2021modeling,chowdhury2022single}, protein generation~\cite{madani2020progen,ferruz2022protgpt2}, and protein-protein interaction prediction~\cite{wang2019high,ofer2021language}. 
To be specific, by treating protein sequences as linguistic strings, these models have demonstrated remarkable effectiveness in predicting protein function based on sequence data alone. 
Technically, as shown in Figure \ref{fig:flow}, protein sequences (e.g., the amino acid sequence \emph{`MLTAHV...'}) are treated as sentences in natural language and amino acids (e.g., `M', `L', and `T') resemble words. 
Thus, Leveraging the powerful BERT architecture originally developed for natural language, ProtBert~\cite{elnaggar2021prottrans} adeptly adapts the BERT~\cite{devlin2018bert} masked language modelling framework to the field of bioinformatics.
% For instance, the protein sequence 'MAWSAVSS......' is treated as sentence, amino acids such as 'M' and 'A' are treated as words. 
This analogy allows ProtBert to employ the technique of predicting randomly masked elements in sequences, thereby learning to identify complex patterns and dependencies among amino acids. 
Similar to ProtBert, ESM~\cite{rives2021biological,verkuil2022language,hie2022high} extends this paradigm by employing a more refined Transformer-based architecture, focusing on capturing the evolutionary relationships and functional dynamics within protein sequences. 
In other words,  most existing protein modelling methods aim to perform protein representation learning by encoding the protein's sequence information for various downstream applications, such as amino acid contact prediction~\cite{singh2022spot}, protein homology detection~\cite{kaminski2023plm}, protein stability prediction~\cite{chu2024protein}, protein-protein interaction identification~\cite{wang2019high,ofer2021language}, etc.

\begin{figure}[t]
    \centering
    \vskip -0.12in
    \subfigure[FfIBP]{\includegraphics[width=0.10\textwidth]{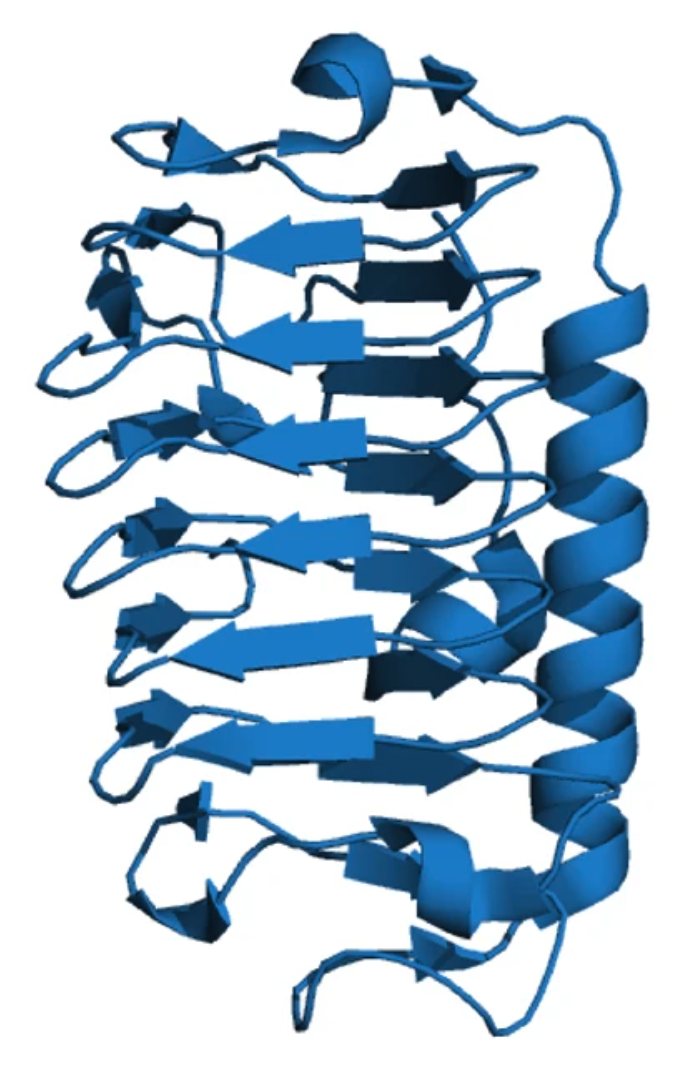}}
    \quad
    \subfigure[CaTrailin\_4]{\includegraphics[width=0.13\textwidth]{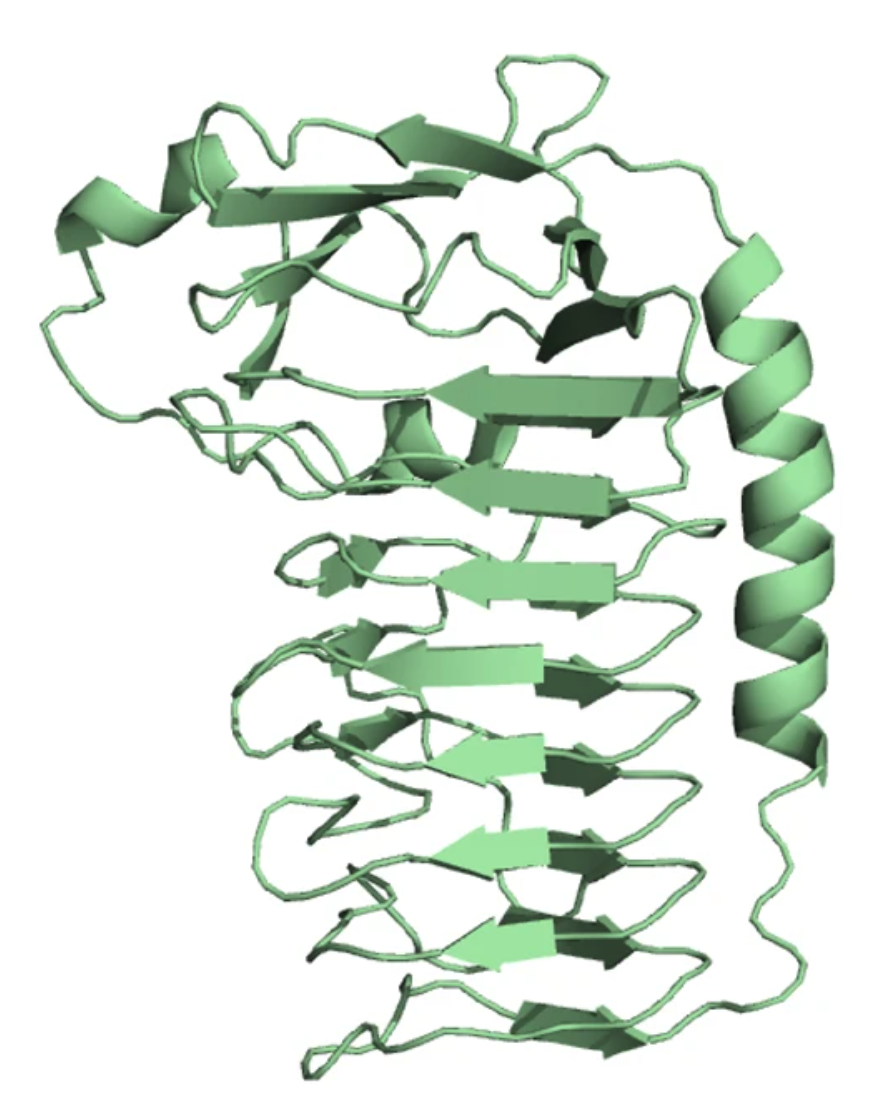}}
    \quad
    \subfigure[Alignment]{\includegraphics[width=0.14\textwidth]{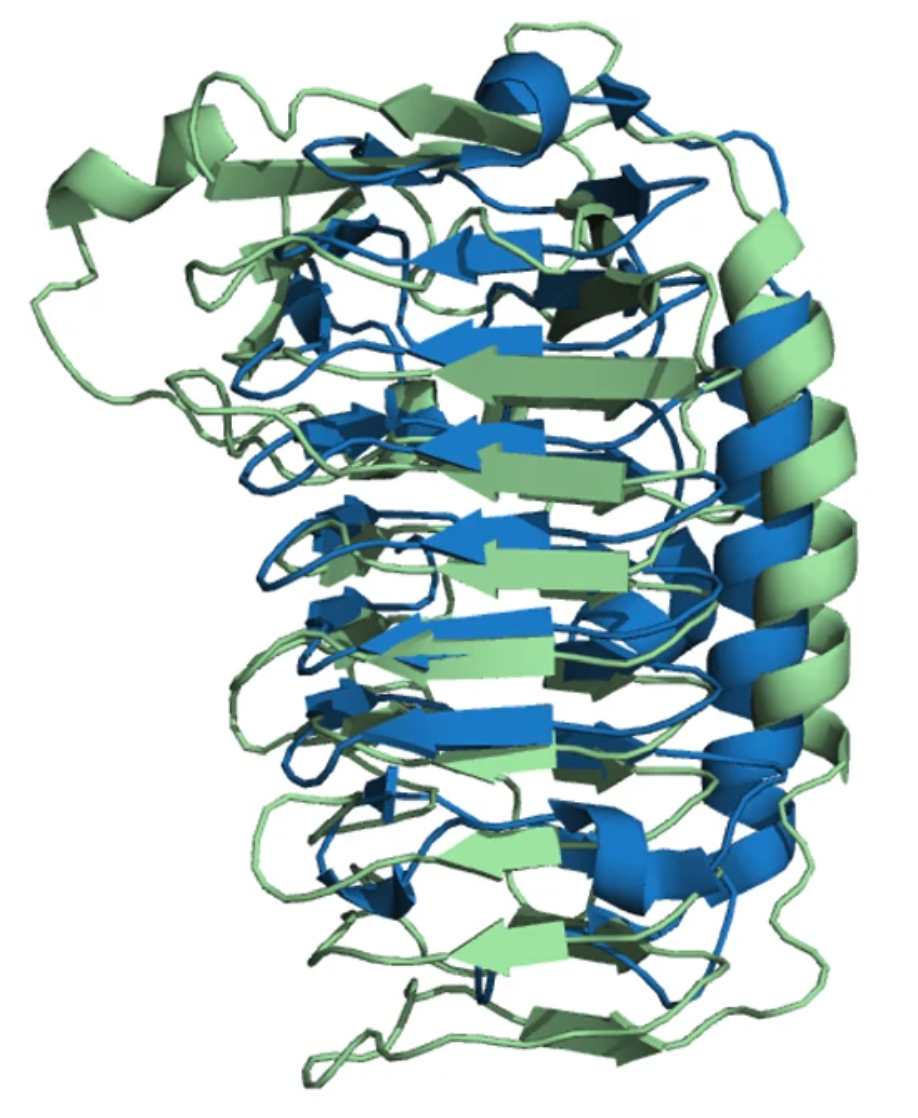}}
    \vskip -0.2in
    \caption{An example of protein structure similarity. 
    Given the predictive structures of a protein pair: (a) the bacterial ice-binding protein FfIBP and  (b) the diatom adhesion protein CaTrailin\_4~\cite{zackova2023diatom,al2023protein}, (c) is FfIBP (blue) and CaTrailin\_4 (green) structure alignment.}
    \label{fig:alignment} 
    \vskip -0.2in
\end{figure}

Despite the aforementioned successes, 
most existing protein language modelling methods suffer from intrinsic limitations.
% \wq{While most of the previous methods have largely adhered to traditional language modelling tasks, most of their focus has primarily been on the linear sequence information of amino acids, often neglecting the crucial aspects of protein structure. }
Specifically, most of their focuses have primarily been on the amino acid sequence, often neglecting the crucial aspects of protein structure. 
Proteins possess the ability to fold into diverse 3D shapes, interacting with various proteins and small molecules in biologically significant ways~\cite{jumper2021highly,mirdita2022colabfold,tsaban2022harnessing}. 
Since \textit{protein's structure determines function}~\cite{greslehner2018molecular},
utilizing protein 3D structure information effectively is crucial for protein language modelling, in which many studies have demonstrated the potential of pre-training on experimentally determined protein structures~\cite{hermosilla2022contrastive,su2023saprot,wang2022lm,zhang2022protein}.
%neverthelness, their utilization of structural information remains constrained.
Nevertheless, these methods focus only on the structure within proteins and ignore the global similarities between proteins. 
We emphasize that the information on protein structure is not only limited to its structure (i.e., conformation) in 3D space but also includes information ranging from local amino acid molecules to the global structural similarity between proteins, as shown in Figure \ref{fig:flow}. 
Local information involves the detailed properties and orientations of individual amino acids, which can affect protein stability and biochemical activity~\cite{renaud2021deeprank}. 
These specifics are vital as they demonstrate how modifications or mutations at the amino acid level can alter the overall structure and functionality of the protein~\cite{jumper2021highly}. 
Furthermore, protein structure similarities provide information on evolutionary relationships and functional classes, which are crucial for understanding how structurally similar proteins of different species can perform similar or complementary functions within biological systems~\cite{hamamsy2023protein}. 
For example, as shown in Figure \ref{fig:alignment}, the bacterial ice-binding protein FfIBP and the diatom adhesion protein CaTrailin\_4 exhibit no detectable sequence similarity despite their functional similarities~\cite{zackova2023diatom,al2023protein}. Their predicted structures exhibit a remarkable similarity (TM-Score = 0.6), with both proteins adopting a beta-helical fold comprised of two units linked by an alpha helix. This structural topology is characteristic of ice-binding proteins. 
Such comparisons are key to predicting the functions of newly discovered proteins based on known structures, thereby enhancing our grasp of complex biological processes and interactions~\cite{lipman1985rapid,hamamsy2022tm,hamamsy2023protein}. 
However, most existing approaches have ineffectively incorporated amino acid molecule information and protein structural similarities into protein representation learning.  

% On the other hand, approaches like ESM-IF~\cite{hsu2022learning} are tailored towards specific protein tasks, such as protein inverse folding, instead of striving for general protein representation. 
% \wq{**shall we have an example of two proteins with similar structures??}
% \wq{However, most existing approaches have ineffectively incorporated amino acid molecule information and protein structural similarities into protein representation learning.  
% }

% \begin{figure}[t]
%     \centering
%     \includegraphics[width=1\linewidth]{figures/results.png}
%     % \vspace{-7mm}
%     \setlength{\belowcaptionskip}{-1cm}
%     \caption{Trasfer learning performance of our GLProtein, SaProt, ProtBert and KeAP on downstream protein analysis tasks. By fine-tuning these pre-trained protein models, GLProtein outperforms others in 9 out of 10 downstream tasks.}
%     \label{fig:results}
% \end{figure}

To eliminate these limitations, we propose a novel protein pre-training framework \textbf{GLProtein} with \textbf{G}lobal-and-\textbf{L}ocal \textbf{Protein} structure information for protein representation learning. Our major contributions are summarized as follows:
\begin{itemize}
    \item We introduce a principled approach for capturing protein structural characteristics in a thorough and detailed manner. This approach incorporates a holistic view of protein structure data, encompassing global structural information, protein structure similarities, as well as local structure information such as protein 3D distance encoding and substructure-based molecular encoding.
    To the best of our knowledge, we are the first to investigate global and local protein structure information in protein language modelling. 
    
    \item We propose a novel protein pre-training framework (GLProtein), where protein structure information is incorporated into protein language models for enhancing protein representation learning.

    \item The comprehensive experiments demonstrate the effectiveness of the proposed method on a wide range of downstream tasks, which verify the performance superiority of GLProtein.
\end{itemize}

% The remainder of this paper is organized as follows. Firstly, we review several works related to our model in Section \ref{sec:relatedwork}. We introduce the proposed framework in Section \ref{sec:methodology}. In Section \ref{sec:Experiments}, we conduct experiments on 8 downstream tasks to illustrate the effectiveness of the proposed method. Finally, we conclude our work with future directions in Section \ref{sec:conclusion}.

% \wq{***we have 3-4 contributions. The 2nd one is not too much useful. 1. about global-local, and 3D structure, 2. the whole framework we proposed, 3. experiments and results!***}

% \begin{figure}[htbp]
%     \centering
%     % \hspace{-10mm}
%     \includegraphics[width=0.5\textwidth]{figures/flow.png}
%     \caption{Protein}
%     \label{fig:protein}
% \end{figure}

% \begin{figure}[htbp]
%     \centering
%     % \hspace{-10mm}
%     \includegraphics[width=0.5\textwidth]{figures/flow.pdf}
%     \caption{flow}
%     \label{fig:flow}
% \end{figure}

\section{Related Work}
\label{sec:relatedwork}
% In this section, we briefly review related work about protein language modelling as well as protein structure modelling. 

\noindent \textbf{Protein Langauge Modelling.} 
As an approach to protein representation learning, protein language modelling is a burgeoning field at the intersection of computational biology and natural language processing (NLP)~\cite{hu2023improving,li2024empowering,li2024molreflect,li2024tomg}. Inspired by the success of language models in NLP, researchers have adapted these techniques to analyse and predict the properties of protein sequences~\cite{fan2025computational}. 
% The earliest approaches to protein language modelling relied on statistical models, such as n-gram models and hidden Markov models (HMMs). These models treated protein sequences as linguistic texts, where amino acids correspond to alphabetic characters in a language ~\cite{durbin1998biological,yoon2009hidden}. The field saw transformative changes with the introduction of neural language models. Early adaptations involved the use of recurrent neural networks (RNNs), particularly LSTM (Long Short-Term Memory) models, to handle the sequential nature of protein data. These models were capable of learning long-range dependencies within sequences, a crucial feature for tasks like epitope mapping and function prediction, such as ProLanGO~\cite{cao2017prolango}, DeepCDA~\cite{abbasi2020deepcda} and DeepSite~\cite{zhang2020deepsite}.
Recent advancements have been dominated by the application of transformer-based models, which utilise self-attention mechanisms to capture relationships between amino acids in a sequence. ProtTrans~\cite{elnaggar2021prottrans} and ESM~\cite{beal2015esm,verkuil2022language,hie2022high}, trained on large-scale protein databases, have shown remarkable ability in tasks such as protein classification and interaction prediction. Moreover, OntoProtein~\cite{zhangontoprotein} and KeAP~\cite{zhou2023protein} incorporated external biological knowledge to enrich protein representations and enhance performance on various downstream tasks. 
However, most of these protein language models do not explicitly consider the spatial structure of proteins and structural similarities between proteins, like our proposed approach.

\begin{figure*}[h]
    \centering
    % \vspace{-1cm}
    \vspace{-0.1in}
    \hspace{5cm}
    \includegraphics[width=1\textwidth]{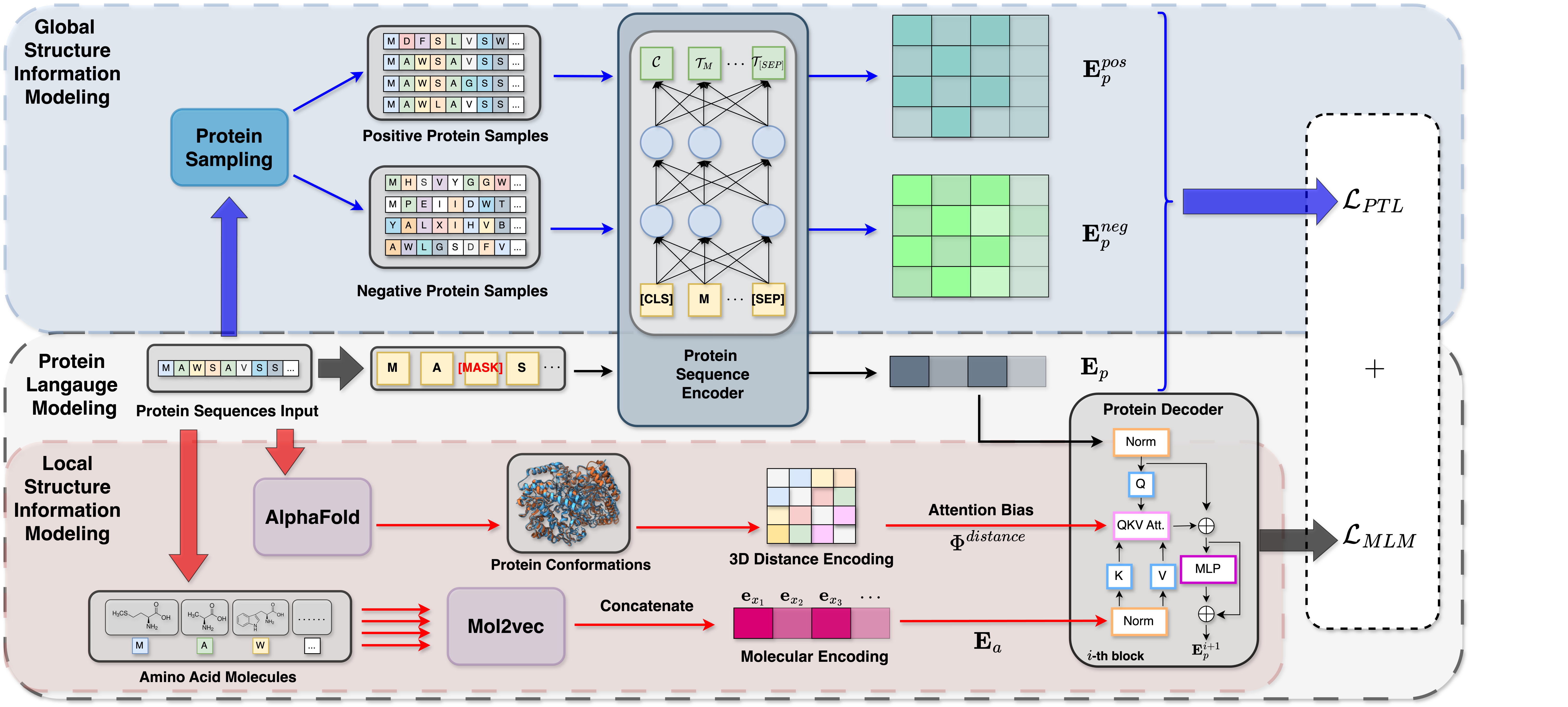}
    \vspace{-0.1in}
    \setlength{\belowcaptionskip}{-3mm}
    \caption{Overview of our proposed model, which jointly optimises global protein similarities and masked protein model with local structure information.}  
    % \vspace{-1cm}
    \label{fig:architecture}  
\end{figure*} 

\noindent \textbf{Protein Structure Modelling.}
The structure of a protein determines its functions. Thus, protein structure modelling has been treated as a reliable way to improve protein representation learning~\cite{huang2024protein,alquraishi2021machine,torrisi2020deep,cheng2008machine}. Some methods use Graph Neural Networks (GNNs) to handle the complex, non-linear relationships inherent in protein structure~\cite{liao2023ld2,jha2022prediction,reau2023deeprank,xu2024deeprank,zhou2023slotgat,zhou2024tined,li2025efficient}. Moreover, RGN2~\cite{chowdhury2022single} utilized a protein language model to learn structural information from unaligned protein sequences. GearNet~\cite{zhang2022protein} focused on geometric pertaining and learned protein features by utilizing spatial relationships between amino acids.   SaProt~\cite{su2023saprot} introduced the concept of a "structure-aware vocabulary" to integrate residue tokens with structure tokens. Similar to the knowledge hancing method, PST~\cite{chen2024endowing} enhances protein language models by integrating structural information through graph transformers to incorporate structural data. Unlike these models, we propose global structure learning and local structure learning methods, which could not only integrate protein structure information and amino acid information but also learn the structure similarity between different proteins by using TM-Score~\cite{hamamsy2023protein}.

\section{Methodology}
\label{sec:methodology}
In this section, we aim to introduce our proposed framework (GLProtein) as a novel solution to learn global and local protein structure information for protein representation learning. 
% We will first give an overview of the proposed framework, then detail each model component, and finally discuss how to learn the model parameters. 
% \subsection{An Overview of the Proposed Framework}
% Most existing protein language modelling methods often have limitations, primarily concentrating on linear sequence information while overlooking the essential aspects of protein structure. 
% To address such limitations, we develop GLProtein, a novel protein pre-training framework that incorporates both global and local protein structure information into protein representation learning. 
We develop GLProtein that incorporates both global and local protein structure information into protein representation learning. 
% More specifically, in order to integrate structure information effectively, a global protein triplet contrastive learning, protein 3D distance encoding, and substructure-based molecular encoding fusion are introduced to empower the protein language model for protein representation learning.  
The framework of GLProtein, shown in Figure \ref{fig:architecture}, consists of three components: \textit{protein language modelling} (Section \ref{protein_language_modelling}), \textit{global structure information modelling} (Section \ref{global}),  and \textit{local structure information modelling} (Section \ref{local}). 
% The first component is protein language modelling, which is the backbone of our proposed framework. It employs random masking to the amino acid sequence and treats local structure information as associated structure information, including 3D distance encoding and substructure-based molecular encoding. 
% The second component is global structure information modelling, as input data in this component is still protein sequence. 
% To obtain a protein triplet, we apply positive and negative protein sampling. 
% After that, we use self-supervised contrastive learning to calculate the protein triplet loss and learn the protein similarity with global structure information by taking advantage of the huge amount of unlabeled data. 
% The third component is local structure information modelling, which integrates local protein structure information in the protein decoder, including 3D distance encoding and substructure-based molecular encoding. Next, we will detail each model component. 

\subsection{Protein Language Modelling}
\label{protein_language_modelling}
% Drawing inspiration from the success of masked language modelling in natural language processing, masked protein modelling conceals part of a protein sequence and tasks a model with predicting the hidden residues~\cite{devlin2018bert,elnaggar2021prottrans}. 
As shown in the center part of Figure ~\ref{fig:architecture}, protein language modelling forms the backbone of our proposed framework, which aims to learn protein representation.
% We adopt a masking strategy similar to KeAP~\cite{zhou2023protein}, where 20\% of the amino acids in a sequence are randomly selected for masking. 
We adopt a masking strategy that each masked amino acid has an 80\% probability of being masked for prediction, a 10\% chance of being replaced with a random amino acid, and a 10\% chance of remaining unchanged. We then integrate protein 3D distance encoding and substructure-based molecular encoding into a protein decoder, in which we will detail in the local structure information modelling component.
Suppose that the number of masked amino acids is $M$ and $x_i$ denotes the $i$-th amino acid. $x_{\sim i}$ denotes the sequence of amino acids excluding the masked amino acid at position $i$. We leverage a cross-entropy loss $\mathcal{L}_{MLM}$ to estimate masked amino acids. Formally, the masked protein modelling objective can be defined as: 
\vskip -0.2in
\begin{small}
    \begin{align}
    \mathcal{L}_{MLM} = -\log \sum_{i \in M}P(x_i | x_{\sim i}; \theta_{E},\theta_{D}),
\end{align}
\end{small}
\vskip -0.1in
where $\theta_{E}$ and $\theta_{D}$ denote the parameters of the protein sequence encoder and decoder, respectively. 
We initialise with a pre-trained BERT-like encoder: ProtBert~\cite{elnaggar2021prottrans}. 
% Next, we will introduce the global structure information modelling part. 

\subsection{Global Structure Information Modelling}
\label{global}

% In protein science, proteins follow the sequence-structure-function paradigm~\cite{koehler2023sequence}, where the amino acid sequence dictates the protein's three-dimensional structure, ultimately determining its function. 
% \textit{Protein's structure determines function},
% Therefore, effectively utilizing information about protein 3D structures is essential for improving protein language modelling. 
% Effectively utilising information about protein 3D structures is essential for improving protein language modelling. 
Protein structures encompass more than mere 3D spatial configurations; they also include global structural information that reflects similarities among proteins. 
% Understanding these structural similarities is crucial, as they can provide insights into evolutionary relationships and functional roles, allowing for more accurate predictions of protein behaviour and interactions~\cite{hamamsy2023protein}.  
To address this, we introduce the concept of global structure information, which contains the structure similarities between proteins, by leveraging the huge amount of self-supervised signals in protein sequences, as shown at the top of Figure~\ref{fig:architecture}. 
To be specific, given each input protein sequence,  positive and negative protein sampling is designed to get the triplet $(P, P_{pos}, P_{neg})$ for capturing protein structure similarity features. 
Then, the protein triplets are encoded to protein representation for the calculation of the contrastive learning loss. 
This optimises the protein sequence encoder by bringing the representation of the input protein $P$ and its positive samples $P_{pos}$ closer together while pushing the representation of $P$ and its negative samples $P_{neg}$ further apart in the representation space.

% \subsubsection{Positive and Negative Protein Sampling}
\paragraph{Positive and Negative Protein Sampling.}

TM-score (Template Modeling Score)~\cite{zhang2004scoring,xu2010significant} is a widely used metric in structural biology for assessing the structural similarity between two protein structures. 
% One of its key advantages is the ability to effectively identify similar three-dimensional structures, even among proteins with low sequence similarity. Additionally, the TM-score is more sensitive to global fold similarity than to local structural variations, allowing it to capture essential information about the overall architecture of proteins. 
% Consequently, 
We utilize the TM-score to measure structural similarity between proteins, focusing on their overall global structure rather than mere sequence identity.
% \wq{The intuition is that proteins with low sequence similarity can still maintain similar three-dimensional structures and thus share functional characteristics. }
Mathematically, the TM-score can be expressed as:
\vskip -0.1in
\begin{small}
    \begin{equation}
    \text{TM-score} = max[\frac{1}{L_N}\sum^{L_r}_{i=1}\frac{1}{1+(\frac{d_i}{d_0})^2}],
\end{equation}
\end{small}
where $L_N$ is the length of the native structure, $L_r$ is the length of the aligned residues to the template structures, $d_i$ is the distance between the $i$-th pair of residues, and $d_0$ is a scaling factor. 

We employ a two-pronged approach that utilizes the TM-Vec model~\cite{hamamsy2023protein} to construct a robust set of positive and negative samples for our protein structure similarity analysis. 
For positive sample selection, we utilize the TM-Vec model to identify the top-K protein sequences that exhibit the highest TM-score values in relation to the template proteins. 
% This approach ensures that our positive samples share significant structural similarity with the template.

In contrast, our negative sampling strategy employs a stochastic selection process followed by structural dissimilarity confirmation. Initially, we randomly select $n$ proteins from our dataset. Subsequently, we employ the TM-Vec model to compute the TM-score between each selected protein and the template protein. Proteins with a TM-score < 0.2 are classified as negative samples, as this threshold indicates a high degree of structural dissimilarity~\cite{xu2010significant}. 
% This method ensures that our negative samples are structurally distinct from the template, providing a clear contrast to the positive samples.

% \subsubsection{Protein Triplet Modelling}
\paragraph{Protein Triplet Modelling.}
After positive and negative protein sampling, we obtain the triplet $(P,P_{pos},P_{neg})$. 
Each protein in the triplet is passed to the protein sequence encoder, resulting in the protein representation, i.e., $\mathbf{E}_p\in \mathbb{R}^{L_p\times D}$, $\mathbf{E}_p^{pos}\in \mathbb{R}^{L_p\times D}$ and $\mathbf{E}_p^{neg}\in \mathbb{R}^{L_p\times D}$. $L_p$ denotes the length of amino acid sequence and $D$ stands for the feature dimension.

% To refine the embedding space by ensuring that proteins with similar structures or functions are clustered closer together than those that are dissimilar, we need to specify an objective function to optimize. 
Since the task we focus on in this part is contrastive learning, the protein triplet loss is designed. This loss function operates by comparing three entities: anchor protein $P$, positive protein $P_{pos}$ and negative protein $P_{neg}$. 
Thus, given protein representation triplet $(P,P_{pos},P_{neg})$, the protein triplet loss $\mathcal{L}_{PTL}$ can be defined as:
\vskip -0.15in
\begin{small}
\begin{equation}
\begin{split}
    &\mathcal{L}_{PTL}(P,P_{pos},P_{neg}) 
    =\\
    &max(|| \mathbf{E}_p - \mathbf{E}^{pos}_{p} ||_2 - || \mathbf{E}_p - \mathbf{E}^{neg}_{p} ||_2 + \epsilon, 0),
\end{split}
\end{equation}
\end{small}
\vskip -0.05in
where $\mathbf{E}_p$, $\mathbf{E}^{pos}_{p}$, $\mathbf{E}^{neg}_{p} \in \mathbb{R}^{L_p\times D}$ are protein representation of the triplet  $(P,P_{pos},P_{neg})$.  $\epsilon$ is a margin between positive and negative pairs.

% \begin{figure*}[htp]
%     \centering
%     \includegraphics[width=1.1\textwidth]{figures/TMVec.pdf}
%     \caption{TMVec.}  
%     \label{fig:TMVec}  
% \end{figure*}

\subsection{Local Structure Information Modelling}
\label{local}
% \wq{Due to the inherent limitations of traditional protein language models that primarily focus on linear amino acid sequences, often neglecting the critical aspects of protein structure. 
% As a part of protein structure information, local protein information encompasses the intricate details of individual amino acids within a protein structure. Thus, we introduce the local structure information modelling component for protein representation learning.}
% Complementing global structural information, local protein information offers additional structure-related insights for protein language modelling. 
While the global structure information modelling component is designed to identify structural similarities across different proteins, the local structure information modelling component zooms in on the specific, intricate features of a protein's internal structure, providing a more nuanced understanding. 
More specifically, in this part, we leverage the local structural details of proteins, including protein 3D distance encoding and substructure-based molecular encoding, to enhance the model's ability to learn and interpret this local configuration effectively, as shown at the bottom of Figure \ref{fig:architecture}. 
% We will first detail the process of protein 3D distance encoding. Then, the integration of embedding of amino acid molecules will be illustrated. 

% \subsubsection{Protein 3D Distance Encoding} 
\paragraph{Protein 3D Distance Encoding.}
% In protein representation learning, the integration of protein 3D coordinates holds immense significance for comprehensively capturing proteins' structural intricacies and functional behaviours. 
The 3D coordinates provide critical insights into how proteins fold and interact in three-dimensional space, influencing their stability, activity, and specificity~\cite{liu2022generating,peng2022pocket2mol,su2023saprot}. 
% By incorporating 3D structural information into representation learning models, the model can capture proteins' geometric and spatial characteristics, which are often overlooked by sequence-based approaches.
We use AlphaFoldDB\footnote{https://alphafold.ebi.ac.uk/} as the 3D protein database and integrate the protein 3D distance encoding~\cite{ying2021transformers} to represent protein 3D structural information to ensure rotational and translational invariance. 
% However, proteins are complex macromolecules composed of numerous amino acids, each consisting of multiple atoms. 
% Consequently, the atomic coordinate data for each protein is extensive.
% To address this challenge, 
We propose to take advantage of the \textbf{alpha-carbon ($\alpha$-C) coordinates} rather than the entire protein coordinates in protein representation learning. 
% That is because the $\alpha$-C atoms are central to the protein backbone and provide a simplified yet informative representation of the protein's 3D structure. 
By capturing the backbone conformation, $\alpha$-C coordinates effectively convey the protein's overall shape and folding pattern, which are critical for understanding its function.
Moreover, leveraging $\alpha$-C coordinates balances capturing essential structural information and maintaining computational efficiency.

Specifically, the coordinates of each $\alpha$-C are processed to represent the current position of the amino acid in 3D space. Then, we encode the Euclidean distance metric to reflect the spatial relation between any pair of amino acids in the 3D space. 
Mathematically, given each amino acid pair $(i,j)$, we first process their Euclidean distance with the Gaussian Basis Kernel function~\cite{scholkopf1997comparing},
$\phi^{k}_{(i,j)} = - \frac{1}{\sqrt{2\pi}|\sigma^k|} \exp(-\frac{1}{2}(\frac{\gamma_{(i,j)}||\mathbf{r_i}-\mathbf{r_j}||+\beta_{(i,j)}-\mu^k}{|\sigma^k|})^2),$
% \begin{small}
%     \begin{equation}
% % \begin{split}
%     \phi^{k}_{(i,j)} = - \frac{1}{\sqrt{2\pi}|\sigma^k|} \exp(-\frac{1}{2}(\frac{\gamma_{(i,j)}||\mathbf{r_i}-\mathbf{r_j}||+\beta_{(i,j)}-\mu^k}{|\sigma^k|})^2),
% % \end{split}
% \end{equation}
% \end{small}

where $k=1, \dots, K$. $K$ is the number of Gaussian Basis kernels. Then, the 3D distance encoding can be calculated as follows:

\begin{small}
    \begin{equation}
% \begin{split}
    \Phi^{ distance}_{(i,j)}=GELU(\bm{\phi}_{(i,j)}\bm{W}^1_D)\bm{W}^2_D,
% \end{split}
\end{equation}
\end{small}
where $\bm{\phi}_{(i,j)} = [\phi^1_{(i,j)};\dots;\phi^K_{(i,j)}]^\top$. $\bm{W}^1_D \in \mathbb{R}^{K\times K}$, $\bm{W}^2_D \in \mathbb{R}^{K\times 1}$ are learnable parameters. $\gamma_{(i,j)}, \beta_{(i,j)}$ are learnable scalars indexed by the pair of amino acid types, and $\mu^k, \sigma^k$ are learnable kernel center and learnable scaling factor of the $k$-th Gaussian Basis Kernel. Denote $\Phi^{ distance}$ as the matrix form of the 3D distance encoding, whose shape is $n \times n$.

% \wq{By incorporating protein coordinate distance encoding, the models can capture the spatial relationships and conformations that are critical to understanding protein functionality and stability. 
% ***We shall put this sentence at the beginning, right?
% }

% \subsubsection{Substructure-based Molecular Encoding}
\paragraph{Substructure-based Molecular Encoding.}
As more detailed information about protein localisation, amino acid molecules play a crucial role in protein representation learning, as they form the essential building blocks of proteins and provide the foundational data for understanding protein structure and function~\cite{lieu2020amino,lopez2024biochemistry}.
To learn the \textbf{fine-grained amino acid structure} information, we introduce substructure-based molecular encoding to leverage the inherent relationships between molecule motifs and substructural features in amino acid molecules.
In practice, we utilize the mol2vec~\cite{jaeger2018mol2vec} to process and derive representations for all amino acid molecules to obtain fine-grained molecular structure information. 
% Then we construct substructure-based molecular encoding as fine-grained protein sequence representation, which is the other part of local structure information modelling component.
For protein $P$, we have 
\vskip -0.2in
\begin{small}
    \begin{equation*}
% \begin{split}
    \mathbf{E}_{a}(P) = \text{Concat}(\mathbf{e}_{x_1},\mathbf{e}_{x_2},\dots,\mathbf{e}_{x_i},\dots,\mathbf{e}_{x_L}), 
% \end{split}    
\end{equation*}
\end{small}
\vskip -0.05in
where $\mathbf{e}_{x_i} \in \mathbb{R}^{1\times d}$, $L$ is the length of the protein sequence, $\mathbf{e}_{x_i}$ is the $i$-th amino acid molecule embedding, and $d$ stands for the feature dimension of the amino acid molecule. 

\subsection{Model Training}
In this part, we will first detail the protein decoder process, which combines protein language modelling and local structure information modelling components. 
Finally, the pre-training objective of the whole framework will be stated. 

% \subsubsection{Protein Decoder}
\paragraph{Protein Decoder.}
% As shown in Figure \ref{fig:architecture}, to enhance the learning of protein representation, we introduce a protein decoder that enables each protein sequence representation to iteratively query and extract relevant information from its associated local structure, including protein 3D distance encoding and substructure-based molecular encoding. This process facilitates the integration of critical structural details, aiding in the accurate reconstruction of masked amino acids.
% Specifically, 
As shown in Figure \ref{fig:architecture}, the decoder treats protein representation $\mathbf{E}_p $ as a query, while the substructure-based molecular encodings $\mathbf{E}_a $ are attended to as keys and values and protein 3D distance encoding $\Phi^{ distance} $ is attended to as attention bias. Taking the $i$-th layer as an example, the inputs to the protein decoder include $\mathbf{E}_{p}^i$, $\Phi^{ distance}$ and $\mathbf{E}_{a}$. The substructure-based molecular encoding $\mathbf{E}_{a}$ is firstly queries by $\mathbf{E}_{p}^i$ as the key and value:
\vskip -0.25in
\begin{small}
    \begin{equation*}
\begin{split}
    Q_p^i = \text{Norm}(\mathbf{E}_{p}^i)W_Q^i,  \\
    K_{a}^i = \text{Norm}(\mathbf{E}_{a})W_K^i, \\
    V_{a}^i = \text{Norm}(\mathbf{E}_{a})W_V^i,
\end{split}        
\end{equation*}
\end{small}
\vskip -0.05in
where $W_Q^i$,$W_K^i$,$W_V^i$ are learnable matrices. Norm stands for the layer normalization~\cite{ba2016layer}.

Then, Attention~\cite{vaswani2017attention} is applied to $\{Q_p^i ,K_{a}^i,V_{a}^i\}$, where the representation of protein sequence extracts helpful, relevant information from the substructure-based molecular encoding. The obtained representation $o_p^i$ stores the helpful structure information for restoring missing amino aids. We then add up $o_p^i$ and $\mathbf{E}_{p}^i$ to integrate information, resulting in the representation $\hat{\mathbf{E}}_p^i$ as follows: 
\vskip -0.20in
\begin{small}
    \begin{equation*}
\begin{split}
    &o_p^i = \text{Attention}(Q_p^i,K_{a}^i,V_{a}^i,\Phi^{ distance}), \\
    &\hat{\mathbf{E}}_p^i = \text{Norm}(\mathbf{E}_p^i) + o_p^i.
\end{split}
\end{equation*}
\end{small}
\vskip -0.1in
% Next, we use $\hat{\mathbf{E}}_p^i$ to query the amino acid molecule term. The whole query, extraction and integration process is similar to that of the coordinate term:
% \begin{equation}
%     \begin{split}
%     &\hat{Q}_p^i = \text{Norm}(\hat{\mathbf{E}}_{p}^i)\hat{W}_Q^i,\\
%     &K_{a}^i = \text{Norm}(\mathbf{E}_{a})\hat{W}_K^i,\\
%     &V_{a}^i = \text{Norm}(\mathbf{E}_{a})\hat{W}_V^i,\\
%     &\hat{o}_p^i = \text{Attention}(\hat{Q}_p^i,K_{a}^i,V_{a}^i),\\
%     &\Bar{\mathbf{E}}_p^i = \text{Norm}(\hat{\mathbf{E}}_p^i) + \hat{o}_p^i.
%     \end{split}
% \end{equation}
The resulting representation $\hat{\mathbf{E}}_p^i$ integrates the helpful, relevant structure information that benefits the restoration of missing amino acids. We finally forward $\hat{\mathbf{E}}_p^i$ through a residual multi-layer perceptron to obtain the output representation of the $i$-th block, which also serves as the input to the $(i+1)$-th block. 

\begin{table*}[h]
\centering
\vspace{-2mm}
\scalebox{0.8} 
{
\begin{tabular}{p{2.0cm}p{1.3cm}p{1.3cm}p{1.3cm}p{1.3cm}p{1.3cm}p{1.3cm}p{1.3cm}p{1.3cm}p{1.3cm}}
\toprule
\multirow{2}{*}{}     & \multicolumn{3}{c}{$6 \le seq < 12$} & \multicolumn{3}{c}{$12 \le seq < 24$} & \multicolumn{3}{c}{$24 \le seq$} \\
\cmidrule(lr){2-4} \cmidrule(lr){5-7} \cmidrule(lr){8-10}
                      & P@L & P@L/2 & P@L/5 & P@L & P@L/2 & P@L/5 & P@L & P@L/2 & P@L/5     \\
\midrule
LSTM       & 0.26$_{(\pm 0.02)}$ & 0.36$_{(\pm 0.01)}$ & 0.49$_{(\pm 0.03)}$ & 0.20$_{(\pm 0.02)}$ & 0.26$_{(\pm 0.02)}$ & 0.34$_{(\pm 0.03)}$ & 0.20$_{(\pm 0.01)}$ & 0.23$_{(\pm 0.02)}$ & 0.27$_{(\pm 0.02)}$ \\

ResNet       & 0.25$_{(\pm 0.02)}$ & 0.34$_{(\pm 0.02)}$ & 0.46$_{(\pm 0.02)}$ & 0.28$_{(\pm 0.01)}$ & 0.25$_{(\pm 0.01)}$ & 0.35$_{(\pm 0.03)}$ & 0.10$_{(\pm 0.03)}$ & 0.13$_{(\pm 0.02)}$ & 0.17$_{(\pm 0.03)}$ \\

Transformer       & 0.28$_{(\pm 0.03)}$ & 0.35$_{(\pm 0.01)}$ & 0.46$_{(\pm 0.02)}$ & 0.19$_{(\pm 0.02)}$ & 0.25$_{(\pm 0.02)}$ & 0.33$_{(\pm 0.01)}$ & 0.17$_{(\pm 0.02)}$ & 0.20$_{(\pm 0.02)}$ & 0.24$_{(\pm 0.02)}$ \\

ProtBert       & 0.30$_{(\pm 0.03)}$ & 0.40$_{(\pm 0.02)}$ & 0.52$_{(\pm 0.02)}$ & 0.27$_{(\pm 0.03)}$ & 0.35$_{(\pm 0.02)}$ & 0.47$_{(\pm 0.01)}$ & 0.20$_{(\pm 0.01)}$ & 0.26$_{(\pm 0.02)}$ & 0.34$_{(\pm 0.01)}$ \\

OntoProtein       & 0.37$_{(\pm 0.02)}$ & 0.46$_{(\pm 0.01)}$ & 0.57$_{(\pm 0.03)}$ & 0.32$_{(\pm 0.01)}$ & 0.40$_{(\pm 0.02)}$ & 0.50$_{(\pm 0.02)}$ & 0.24$_{(\pm 0.03)}$ & 0.31$_{(\pm 0.01)}$ & 0.39$_{(\pm 0.03)}$ \\

LM-GVP    & 0.35$_{(\pm 0.02)}$ & 0.42$_{(\pm 0.02)}$ & 0.49$_{(\pm 0.02)}$ & 0.33$_{(\pm 0.03)}$ & 0.43$_{(\pm 0.02)}$ & 0.51$_{(\pm 0.03)}$ & 0.26$_{(\pm 0.02)}$ & 0.37$_{(\pm 0.02)}$ & 0.43$_{(\pm 0.03)}$ \\ 

GearNet    & 0.39$_{(\pm 0.02)}$ & 0.46$_{(\pm 0.02)}$ & 0.57$_{(\pm 0.02)}$ & 0.36$_{(\pm 0.03)}$ & 0.44$_{(\pm 0.02)}$ & 0.55$_{(\pm 0.03)}$ & 0.29$_{(\pm 0.02)}$ & 0.37$_{(\pm 0.01)}$ & 0.45$_{(\pm 0.02)}$ \\ 

SaProt &0.41$_{(\pm 0.02)}$ & 0.39$_{(\pm 0.03)}$ & 0.42$_{(\pm 0.02)}$ & 0.38$_{(\pm 0.01)}$ & 0.37$_{(\pm 0.01)}$ & 0.41$_{(\pm 0.01)}$ & 0.24$_{(\pm 0.02)}$ & 0.27$_{(\pm 0.03)}$ & 0.37$_{(\pm 0.02)}$\\

KeAP       & 0.41$_{(\pm 0.04)}$ & 0.52$_{(\pm 0.02)}$ & 0.62$_{(\pm 0.03)}$ & 0.36$_{(\pm 0.01)}$ & 0.45$_{(\pm 0.01)}$ & 0.57$_{(\pm 0.01)}$ & 0.29$_{(\pm 0.02)}$ & 0.37$_{(\pm 0.03)}$ & 0.46$_{(\pm 0.02)}$ \\

ESM-2       & 0.42$_{(\pm 0.02)}$ & 0.49$_{(\pm 0.03)}$ & 0.63$_{(\pm 0.01)}$ & 0.37$_{(\pm 0.01)}$ & 0.43$_{(\pm 0.01)}$ & 0.57$_{(\pm 0.02)}$ & 0.30$_{(\pm 0.02)}$ & 0.38$_{(\pm 0.03)}$ & 0.46$_{(\pm 0.02)}$ \\
\rowcolor{gray!20}
\textbf{GLProtein}       & \textbf{0.45}$_{(\pm 0.02)}$ & \textbf{0.55}$_{(\pm 0.02)}$ & \textbf{0.66}$_{(\pm 0.01)}$ & \textbf{0.39}$_{(\pm 0.03)}$ & \textbf{0.48}$_{(\pm 0.01)}$ & \textbf{0.58}$_{(\pm 0.02)}$ & \textbf{0.31}$_{(\pm 0.02)}$ & \textbf{0.40}$_{(\pm 0.01)}$ & \textbf{0.47}$_{(\pm 0.02)}$ \\
\bottomrule
\end{tabular}
}
\caption{
Comparisons on amino acid contact prediction. \textbf{seq} signifies the distance, measured in terms of amino acid units, between two selected amino acids. \textbf{P@L, P@L/2, P@L/5} denote the precision scores calculated upon top L (i.e., L most likely contacts), top L/2, and top L/5 predictions, respectively. The best results are bolded, and the second-best results are underlined.}
\label{Contact Prediction}
\end{table*}

\begin{figure*}[h]
	
	% \begin{minipage}{0.24\linewidth}
	% 	\vspace{3pt}
 %        %这个图片路径替换成你的图片路径即可使用
	% 	\centerline{\includegraphics[width=\textwidth]{figures/label.png}}
 %          % 加入对这列的图片说明
	% 	\centerline{Labels}
	% \end{minipage}
	% \begin{minipage}{0.24\linewidth}
	% 	\vspace{3pt}
	% 	\centerline{\includegraphics[width=\textwidth]{figures/GLProtein.png}}
	 
	% 	\centerline{GLProtein}
	% \end{minipage}
	% \begin{minipage}{0.24\linewidth}
	% 	\vspace{3pt}
	% 	\centerline{\includegraphics[width=\textwidth]{figures/KeAP.png}}
	 
	% 	\centerline{KeAP}
	% \end{minipage}
 %        \begin{minipage}{0.24\linewidth}
	% 	\vspace{3pt}
	% 	\centerline{\includegraphics[width=\textwidth]{figures/protBERT.png}}
	 
	% 	\centerline{ProtBert}
	% \end{minipage}

 %        \qquad
	% %让图片换行，

 %        \begin{minipage}{0.20\linewidth}
	% 	\vspace{3pt}
	% 	\centerline{\includegraphics[width=\textwidth]{figures/TBM.png}}
	 
	% 	\centerline{ProtBert}
	% \end{minipage}
        
        \begin{minipage}{1.0\linewidth}
		\vspace{3pt}
        %这个图片路径替换成你的图片路径即可使用
		\centerline{\includegraphics[width=\textwidth]{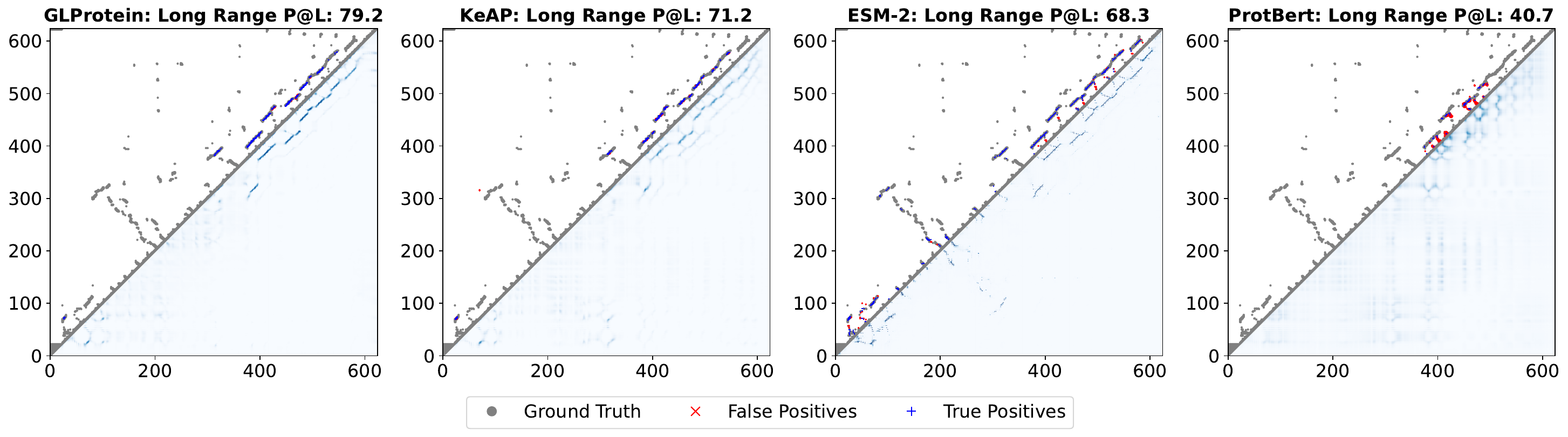}}
          % 加入对这列的图片说明
		\centerline{}
	\end{minipage}

	% \caption{Visualization of contact prediction comparison. The leftmost one is the amino acid contact map of a randomly selected protein, and the others are contact probability maps of contact predictions by GLProtein, KeAP and ProtBert.}
        \vskip -0.3in
        \caption{An example of amino acid contacts (top-L predictions for ProteinNet~\cite{alquraishi2019proteinnet} test example \text{TBM-hard\#T0912}). Raw contact probabilities are shown below the diagonal, top L contacts are shown above the diagonal (blue: true positives, red: false positives, grey: ground-truth contacts).}
	\label{contact_examples}
\end{figure*}

% By training with both protein 3D distance encoding and substructure-based molecular encoding, the model can not only go beyond simply memorising sequence patterns but also actively decode the underlying structural and chemical principles that govern protein functionality. 

% \subsubsection{Pre-training Objective}
\paragraph{Pre-training Objective.}
To estimate the model parameters of GLProtein, we adopt the masked protein modelling object and global protein triplet objective to construct the overall model. We jointly optimize the overall objective as follows:

\begin{small}
    \begin{equation}
    \mathcal{L} = \mathcal{L}_{MLM} + \alpha \mathcal{L}_{PTL},
\end{equation}
\end{small}
where $\alpha$ is the hyper-parameter.

\begin{figure*}[htbp]
    \centering
    \includegraphics[width=1\linewidth]{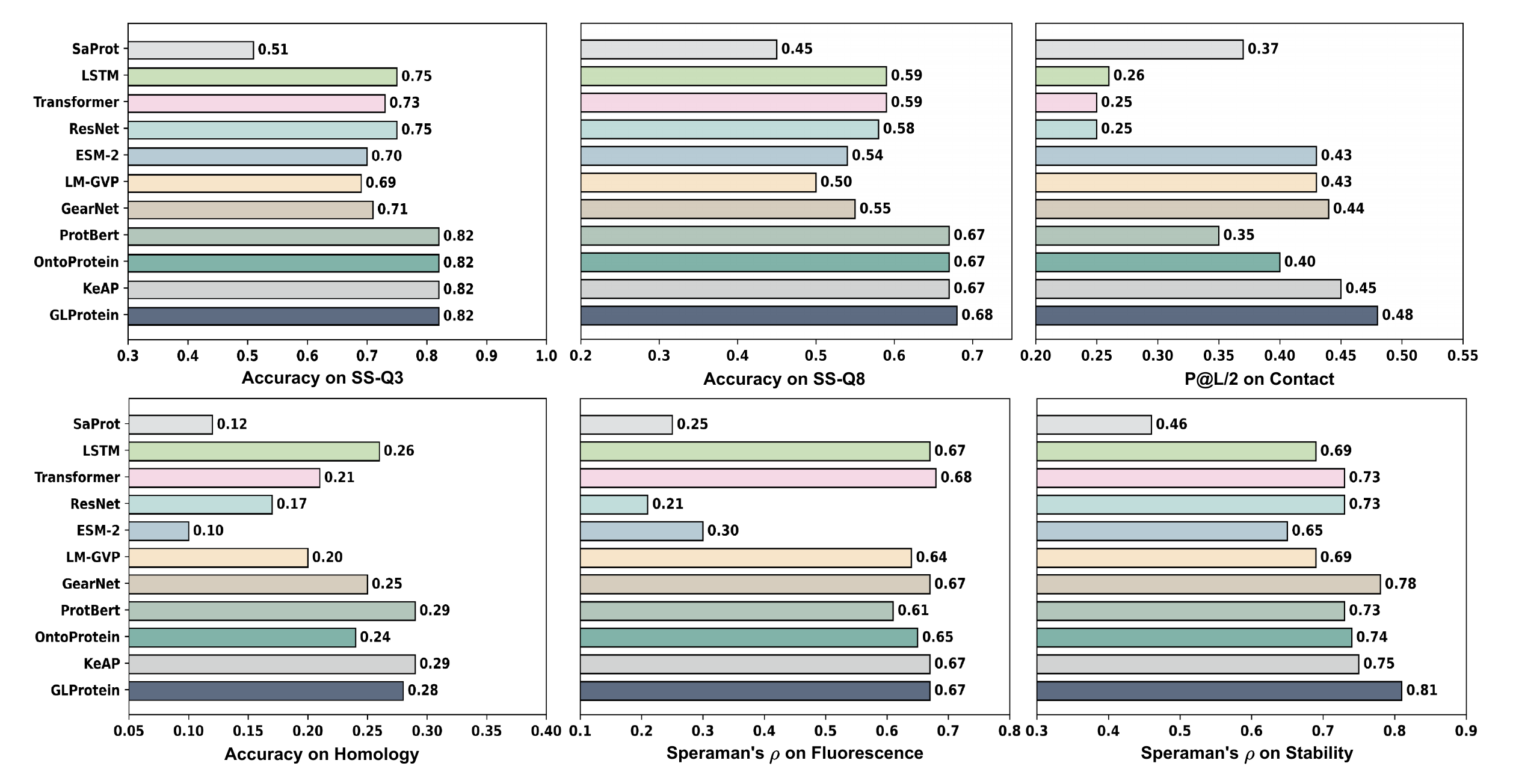}
    % \vspace{-5mm}
    \vskip -0.2in
    \caption{Results on TAPE Benchmark encompass various evaluations. SS is a secondary structure task that is evaluated in CB315. We report medium- and long-range results using P@L/2 metrics in contact prediction task. In fluorescence and stability prediction tasks, we use Spearman's $\rho$ metric for evaluation. We also provide a related table in Appendix \ref{tape_results}.}
    \label{fig:tape}
\end{figure*}

\section{Experiments}
\label{sec:Experiments}

In this section, we evaluate the generalization ability of the learned protein representation by fine-tuning the pre-trained model across a diverse array of downstream applications, including amino acid contact prediction, protein homology detection, protein stability prediction, protein-protein interaction identification, protein-protein binding affinity prediction and semantic similarity inference. 

% \subsection{Pre-training Dataset and Implementation Details}
% ProteinKG25 ~\cite{zhangontoprotein} offers a comprehensive knowledge graph comprising approximately five million triplets. This graph includes nearly 600k proteins, 50k attribute terms and 31 relation terms. However, note that we only use the protein sequence data in this dataset. Our model does not use the attribute and relation terms extracted from Gene Ontology. We used modified pre-training data from KeAP~\cite{zhou2023protein}, which removed overlapping data between ProteinKG25 and downstream datasets. 
\paragraph{Pretraining Datasets.}
Swiss-Prot~\cite{boeckmann2003swiss} offers a comprehensive and manually curated protein sequence database that includes nearly 600k protein sequences. We use it as pertaining dataset. 
Additionally, we use AlphaFoldDB to obtain the protein 3D coordinate datasets. 

\paragraph{Implementation Details.}
We conducted some experiments and compared GLProtein with baselines regarding pre-training and inference time in contact prediction tasks, as shown in Appendix Table \ref{time}. GLProtein outperforms baselines in multiple downstream tasks with similar parameters. During pre-training, GLProtein is trained for 300k steps using a learning rate of 1e-5, weight decay of 0.01 over four GPUs (NVIDIA A6000, 48G Memory each). We uniformly fine-tuned all downstream tasks without structural information to ensure fair and unbiased comparisons. For the amino acid contact prediction and protein-protein interaction task, we randomly selected five random seeds to fine-tune our model and the baseline model separately and report the results.

\subsection{Downstream Tasks}
\subsubsection{Amino Acid Contact Prediction}\

% \rl{When presenting your results, could use a **Improvement** Line to show how much you improve (some improvements are amazing)}
% \paragraph{Amino acid contact prediction.}
\noindent \textbf{Overview.} 
Amino acid contact prediction is a critical task in computational biology, aiming to identify pairs of amino acids within a protein that are in close spatial proximity. Given an input protein sequence, our model predicts whether pairs of amino acids from the same sequence are in contact. The model accomplishes this by generating a probability contact matrix for each input protein. We tested the model on the dataset collected and organized by ProteinNet~\cite{alquraishi2019proteinnet} and TAPE~\cite{rao2019evaluating}.

\noindent \textbf{Baselines.} We evaluate our model compared with ten baselines. Specifically, we employed variations of LSTM~\cite{hochreiter1997long}, ResNet~\cite{he2016deep} and Transformer~\cite{vaswani2017attention} proposed by the TAPE benchmark~\cite{rao2019evaluating}. ProtBert~\cite{elnaggar2021prottrans} is a BERT-like model pre-trained on UniRef100~\cite{suzek2007uniref,suzek2015uniref}. ESM-2~\cite{rives2021biological,verkuil2022language,hie2022high} feature a transformer architecture pre-trained on the representative sequences from UniRef50~\cite{suzek2007uniref,suzek2015uniref}. OntoProtein~\cite{zhangontoprotein} and KeAP~\cite{zhou2023protein} are the most recent knowledge-based pre-training methodologies. SaProt~\cite{su2023saprot} is the most recent structure-based protein language model. LM-GVP~\cite{wang2022lm} and GearNet~\cite{zhang2022protein} are famous geometric methods for protein representation learning. We uniformly fine-tuned all downstream tasks without structural information to ensure fair and unbiased comparisons. Since all structural tokens are masked, residual information still exists; we substitute Foldseek structure tokens with "\#" when fine-tuning SaProt. 

\noindent \textbf{Results.} Table ~\ref{Contact Prediction} shows the experimental results of amino acid contact prediction. Specifically, we notice that our model GLProtein consistently outperforms other models in short- ($6 \leq seq < 12$), medium- ($12 \leq 
 seq < 24$) and long-range ($seq > 24$) contact predictions. 
Notably, our model demonstrates better performance compared to SaProt, which is also a structure-based language model. We also randomly selected a protein from the contact test dataset for visual analysis. As shown in Figure \ref{contact_examples}, the left is our GLProtein's result of amino acid contacts. The right three are the contact maps of three baseline models, including KeAP, ESM-2 and ProtBERT. Figure \ref{contact_examples} shows more visually that GLProtein's prediction on the task of contact prediction is closer to labels, i.e., better performance on long-range contact prediction. We attribute the enhancements in performance achieved by GLProtein to its innovative integration of global and local structural information, which allows the pre-trained model to gain a deeper understanding of protein structure. More results can be found in the Appendix \ref{tape_results}.

\subsubsection{Protein-Protein Interaction}\

\noindent \textbf{Overview.} Protein-protein interaction (PPI) is fundamental to virtually all biological processes and pathways in living organisms. It refers to the physical contact between two or more amino acid sequences. In this paper, we only focus on two-protein cases where a pair of protein sequences serve as the inputs. The objective is to accurately predict the specific types of interactions that occur between each pair of proteins.

In our experiments, we focus on predicting 7 interaction types between protein pairs, namely reaction, binding, post-translational modifications, activation, inhibition, catalysis, and expression. The challenge of PPI prediction is approached as a multi-label classification problem. We conducted our experiments using three datasets: SHS27K~\cite{chen2019multifaceted}, SHS148K~\cite{chen2019multifaceted} and STRING~\cite{lv2021learning}. Both SHS27K and SHS148K are considered subsets of STRING, with proteins excluded if they have fewer than 50 amino acids or exhibit 40\% or higher sequence identity. We followed OntoProtein's setting to generate test sets and employed Breadth-First Search (BFS) and Depth-First Search (DFS) techniques across these datasets. The F1 score is utilized as the primary metric for evaluating performance.

\noindent  \textbf{Baselines.} Following OntoProtein~\cite{zhangontoprotein} and KeAP\cite{zhou2023protein}, we have expanded our baseline models to include four additional methods: DPPI~\cite{hashemifar2018predicting}, DNN-PPI~\cite{li2018deep}, PIPR~\cite{chen2019multifaceted} and GNN-PPI~\cite{lv2021learning}. These are incorporated alongside existing baselines such as ProtBert~\cite{elnaggar2021prottrans}, ESM-2~\cite{beal2015esm}, OntoProtein~\cite{zhangontoprotein}, KeAP~\cite{zhou2023protein}, SaProt~\cite{su2023saprot}, LM-GVP~\cite{wang2022lm}, GearNet~\cite{zhang2022protein}, DeepInter~\cite{lin2023protein}, MAPE-PPI~\cite{wu2024mape}, ProLLM~\cite{jin2024prollm} and ESM-C~\cite{hayes2025simulating}, providing a comprehensive set of comparisons in our analysis. 

\begin{table}[htp]

    \centering
    \vspace{-0.1in}
    \resizebox{\linewidth}{!}{ 
    \begin{tabular}{l cccccc}
    \toprule
    \multicolumn{1}{l}{} &
    \multicolumn{2}{c}{\textbf{SHS27k}} &
    \multicolumn{2}{c}{\textbf{SHS148k}} &
    \multicolumn{2}{c}{\textbf{STRING}} \\
    \cmidrule(lr){2-3} \cmidrule(lr){4-5} \cmidrule(lr){6-7}
    \textbf{Methods} & BFS & DFS & BFS & DFS & BFS & DFS \\
    \midrule
     DPPI  & 40.27$_{(\pm 0.74)}$ & 44.86$_{(\pm 0.87)}$ & 51.26$_{(\pm 0.66)}$ &  51.43$_{(\pm 0.94)}$ &  55.79$_{(\pm 0.81)}$ &  64.72$_{(\pm 0.94)}$ \\
      DNN-PPI &   47.97$_{(\pm 0.94)}$ &   52.85$_{(\pm 0.91)}$ &   55.90$_{(\pm 0.67)}$ &   57.82$_{(\pm 0.78)}$ &   52.74$_{(\pm 0.89)}$  &   62.99$_{(\pm 0.93)}$ \\
    PIPR & 43.67$_{(\pm 0.99)}$ & 56.76$_{(\pm 0.82)}$ & 60.10$_{(\pm 0.85)}$ & 61.83$_{(\pm 0.94)}$ & 53.65$_{(\pm 0.88)}$ & 66.46$_{(\pm 0.92)}$ \\
    GNN-PPI & 62.47$_{(\pm 0.65)}$ & 73.19$_{(\pm 0.89)}$ & 71.01$_{(\pm 0.92)}$ & 81.54$_{(\pm 0.87)}$ & 75.34$_{(\pm 0.82)}$ & 90.01$_{(\pm 0.78)}$ \\
    ProtBert & 68.44$_{(\pm 0.78)}$ & 72.36$_{(\pm 0.85)}$ & 70.06$_{(\pm 0.88)}$ & 77.46$_{(\pm 0.62)}$ & 66.08$_{(\pm 0.91)}$ & 86.45$_{(\pm 0.82)}$ \\
    DeepInter     & 77.31$_{(\pm 1.14)}$ & 77.18$_{(\pm 0.84)}$ & 74.52$_{(\pm 0.99)}$ & 76.60$_{(\pm 0.49)}$ & 77.82$_{(\pm 0.98)}$ & 80.04$_{(\pm 1.18)}$ \\
      OntoProtein &   71.37$_{(\pm 0.84)}$ &   76.28$_{(\pm 0.77)}$ &    74.60$_{(\pm 0.56)}$ &  76.33$_{(\pm 0.69)}$  &    75.64$_{(\pm 0.91)}$  &   90.23$_{(\pm 0.79)}$ \\
      KeAP &   78.51$_{(\pm 0.95)}$ &   78.84$_{(\pm 0.85)}$ &    74.26$_{(\pm 0.89)}$ &  81.99$_{(\pm 0.92)}$ &    80.08$_{(\pm 0.79)}$  &   88.47$_{(\pm 0.71)}$ \\
      LM-GVP &   80.25$_{(\pm 1.24)}$ &  79.4$2_{(\pm 0.83)}$ &    77.6$_{(\pm 0.76)}$ &  80.36$_{(\pm 0.97)}$ &    81.17$_{(\pm 0.58)}$  &   85.67$_{(\pm 0.74)}$\\
      MAPE-PPI     & 83.63$_{(\pm 0.76)}$  & 81.01$_{(\pm 0.58)}$  & 84.57$_{(\pm 0.91)}$  & 83.62$_{(\pm 0.69)}$  & 87.18$_{(\pm 0.82)}$  & 87.46$_{(\pm 0.59)}$ \\
      GearNet &   85.46$_{(\pm 0.61)}$ &   82.73$_{(\pm 0.69)}$ &    80.02$_{(\pm 1.26)}$ &  82.28$_{(\pm 0.93)}$ &    85.55$_{(\pm 0.74)}$  &   88.03$_{(\pm 0.51)}$\\
      ESM-2 &   94.01$_{(\pm 0.77)}$ &   87.32$_{(\pm 0.97)}$ &    91.46$_{(\pm 0.63)}$ &  85.24$_{(\pm 0.46)}$ &    88.13$_{(\pm 0.71)}$  &   85.53$_{(\pm 0.55)}$ \\
      SaProt &  91.18$_{(\pm 0.73)}$ &   88.85$_{(\pm 1.04)}$ &    90.75$_{(\pm 0.91)}$ &  80.67$_{(\pm 0.90)}$ &   88.23$_{(\pm 0.81)}$ &   88.90$_{(\pm 0.74)}$ \\
      ProLLM & 91.49$_{(\pm 0.91)}$ & 88.38$_{(\pm 0.78)}$ & 90.90$_{(\pm 1.03)}$ & 85.34$_{(\pm 0.61)}$ & 87.38$_{(\pm 0.77)}$ & 86.99$_{(\pm 0.57)}$ \\
      ESM-C & 92.46$_{(\pm 1.02)}$ & 88.14$_{(\pm 0.75)}$ & 91.86$_{(\pm 0.51)}$ & 86.11$_{(\pm 0.78)}$ & 87.41$_{(\pm 0.95)}$ & 86.78$_{(\pm 0.66)}$ \\
      \rowcolor{gray!20}
      \textbf{GLProtein} &  \textbf{96.32}$_{(\pm 0.86)}$ &  \textbf{91.23}$_{(\pm 0.92)}$ &   \textbf{93.78}$_{(\pm 0.77)}$ &  \textbf{86.14}$_{(\pm 0.69)}$ &    \textbf{89.41}$_{(\pm 0.66)}$  &  \textbf{91.35}$_{(\pm 0.89)}$\\

    \bottomrule
    \end{tabular}}
    \caption{
    Protein-Protein Interaction Prediction Results.
    Breath-First Search (BFS) and Depth-First Search (DFS) are strategies that split the training and testing PPI datasets. The best results are bolded, and the second-best results are underlined.
    }
    \label{tab:ppi_result}
\end{table}

 \noindent \textbf{Results.} As shown in Table \ref{tab:ppi_result}, the results clearly indicate that our method consistently outperforms all other methods, including the structure-based protein language model such as SaProt and multimodal protein language model such as ESM-C, across all datasets and both BFS and DFS evaluation metrics. The observed decline in performance can be linked to the growing amount of fine-tuning data, transitioning from SHS27k to STRING, which diminished the influence of pre-training. We believe that the structural similarities between proteins identified during the pre-training step enable GLProtein to excel in the PPI task, resulting in its outstanding performance.

\section{Conclusion and Future Work}
\label{sec:conclusion}

In this work, we propose GLProtein, a general protein language model with global and local protein structure information. GLProtien outperforms the previous protein representation learning model on most downstream applications, demonstrating the performance superiority of GLProtein. In the future, we aim to further enhance GLProtein's capabilities by exploring novel avenues for incorporating multi-modal data sources, refining the model's interpretability, and extending its applicability to a wider array of biological contexts. 

\section{Limitations}
We have observed that GLProtein underperforms on certain individual tasks. For instance, in the protein-protein binding affinity prediction task, ESM-2 surpasses GLProtein. This task focuses on predicting changes in binding affinity resulting from protein mutations. We believe that GLProtein's limited performance is attributed to its lack of mutation information, whereas ESM-2 incorporates multiple sequence alignment (MSA) data during training, which includes mutation insights.
Similarly, in the Fluorescence task, GLProtein does not demonstrate significant improvement when tasked with distinguishing closely related proteins. We hypothesize that while GLProtein effectively learns structural similarities among different proteins during pre-training, it excels at identifying differences between dissimilar structures but struggles to differentiate between similar ones. We plan to further investigate these issues in our future research. 

\section{Acknowledgements}

The research described in this paper has been partially supported by the National Natural Science Foundation of China (project no. 62102335 and 62372314), General Research Funds from the Hong Kong Research Grants Council (project no. PolyU 15207322, 15200023, 15206024, and 15224524), internal research funds from Hong Kong Polytechnic University (project no. P0042693, P0048625, and P0051361). This work was supported by computational resources provided by The Centre for Large AI Models (CLAIM) of The Hong Kong Polytechnic University.

% Bibliography entries for the entire Anthology, followed by custom entries
%\bibliography{anthology,custom}
% Custom bibliography entries only
\bibliography{custom}

\begin{thebibliography}{88}
\providecommand{\natexlab}[1]{#1}

\bibitem[{Al-Fatlawi et~al.(2023)Al-Fatlawi, Menzel, and Schroeder}]{al2023protein}
Ali Al-Fatlawi, Martin Menzel, and Michael Schroeder. 2023.
\newblock Is protein blast a thing of the past?
\newblock \emph{nature communications}, 14(1):8195.

\bibitem[{AlQuraishi(2019)}]{alquraishi2019proteinnet}
Mohammed AlQuraishi. 2019.
\newblock Proteinnet: a standardized data set for machine learning of protein structure.
\newblock \emph{BMC bioinformatics}, 20:1--10.

\bibitem[{AlQuraishi(2021)}]{alquraishi2021machine}
Mohammed AlQuraishi. 2021.
\newblock Machine learning in protein structure prediction.
\newblock \emph{Current opinion in chemical biology}, 65:1--8.

\bibitem[{Ba et~al.(2016)Ba, Kiros, and Hinton}]{ba2016layer}
Jimmy~Lei Ba, Jamie~Ryan Kiros, and Geoffrey~E Hinton. 2016.
\newblock Layer normalization.
\newblock \emph{arXiv preprint arXiv:1607.06450}.

\bibitem[{Beal(2015)}]{beal2015esm}
Daniel~J Beal. 2015.
\newblock Esm 2.0: State of the art and future potential of experience sampling methods in organizational research.
\newblock \emph{Annu. Rev. Organ. Psychol. Organ. Behav.}, 2(1):383--407.

\bibitem[{Boeckmann et~al.(2003)Boeckmann, Bairoch, Apweiler, Blatter, Estreicher, Gasteiger, Martin, Michoud, O'Donovan, Phan et~al.}]{boeckmann2003swiss}
Brigitte Boeckmann, Amos Bairoch, Rolf Apweiler, Marie-Claude Blatter, Anne Estreicher, Elisabeth Gasteiger, Maria~J Martin, Karine Michoud, Claire O'Donovan, Isabelle Phan, and 1 others. 2003.
\newblock The swiss-prot protein knowledgebase and its supplement trembl in 2003.
\newblock \emph{Nucleic acids research}, 31(1):365--370.

\bibitem[{Chandonia et~al.(2019)Chandonia, Fox, and Brenner}]{chandonia2019scope}
John-Marc Chandonia, Naomi~K Fox, and Steven~E Brenner. 2019.
\newblock Scope: classification of large macromolecular structures in the structural classification of proteins—extended database.
\newblock \emph{Nucleic acids research}, 47(D1):D475--D481.

\bibitem[{Chen et~al.(2024)Chen, Hartout, Pellizzoni, Oliver, and Borgwardt}]{chen2024endowing}
Dexiong Chen, Philip Hartout, Paolo Pellizzoni, Carlos Oliver, and Karsten Borgwardt. 2024.
\newblock Endowing protein language models with structural knowledge.
\newblock \emph{arXiv preprint arXiv:2401.14819}.

\bibitem[{Chen et~al.(2019)Chen, Ju, Zhou, Chen, Zhang, Chang, Zaniolo, and Wang}]{chen2019multifaceted}
Muhao Chen, Chelsea J-T Ju, Guangyu Zhou, Xuelu Chen, Tianran Zhang, Kai-Wei Chang, Carlo Zaniolo, and Wei Wang. 2019.
\newblock Multifaceted protein--protein interaction prediction based on siamese residual rcnn.
\newblock \emph{Bioinformatics}, 35(14):i305--i314.

\bibitem[{Cheng et~al.(2008)Cheng, Tegge, and Baldi}]{cheng2008machine}
Jianlin Cheng, Allison~N Tegge, and Pierre Baldi. 2008.
\newblock Machine learning methods for protein structure prediction.
\newblock \emph{IEEE reviews in biomedical engineering}, 1:41--49.

\bibitem[{Chowdhury et~al.(2022)Chowdhury, Bouatta, Biswas, Floristean, Kharkar, Roy, Rochereau, Ahdritz, Zhang, Church et~al.}]{chowdhury2022single}
Ratul Chowdhury, Nazim Bouatta, Surojit Biswas, Christina Floristean, Anant Kharkar, Koushik Roy, Charlotte Rochereau, Gustaf Ahdritz, Joanna Zhang, George~M Church, and 1 others. 2022.
\newblock Single-sequence protein structure prediction using a language model and deep learning.
\newblock \emph{Nature Biotechnology}, 40(11):1617--1623.

\bibitem[{Chu et~al.(2024)Chu, Narang, and Siegel}]{chu2024protein}
Simon~KS Chu, Kush Narang, and Justin~B Siegel. 2024.
\newblock Protein stability prediction by fine-tuning a protein language model on a mega-scale dataset.
\newblock \emph{PLOS Computational Biology}, 20(7):e1012248.

\bibitem[{Cuff and Barton(1999)}]{cuff1999evaluation}
James~A Cuff and Geoffrey~J Barton. 1999.
\newblock Evaluation and improvement of multiple sequence methods for protein secondary structure prediction.
\newblock \emph{Proteins: Structure, Function, and Bioinformatics}, 34(4):508--519.

\bibitem[{Davis et~al.(2024)Davis, Alexander, Moreno-Cruz, Hong, Shaner, Caldeira, and McKay}]{davis2024food}
Steven~J Davis, Kathleen Alexander, Juan Moreno-Cruz, Chaopeng Hong, Matthew Shaner, Ken Caldeira, and Ian McKay. 2024.
\newblock Food without agriculture.
\newblock \emph{Nature Sustainability}, 7(1):90--95.

\bibitem[{Devlin et~al.(2018)Devlin, Chang, Lee, and Toutanova}]{devlin2018bert}
Jacob Devlin, Ming-Wei Chang, Kenton Lee, and Kristina Toutanova. 2018.
\newblock Bert: Pre-training of deep bidirectional transformers for language understanding.
\newblock \emph{arXiv preprint arXiv:1810.04805}.

\bibitem[{Ding et~al.(2019)Ding, Weng, Liu, Song, Yin, Yuan, Ren, Lei, and Chiang}]{ding2019selective}
Bo~Ding, Yue Weng, Yunqing Liu, Chunlan Song, Le~Yin, Jiafan Yuan, Yanrui Ren, Aiwen Lei, and Chien-Wei Chiang. 2019.
\newblock Selective photoredox trifluoromethylation of tryptophan-containing peptides.
\newblock \emph{European Journal of Organic Chemistry}, 2019(46):7596--7605.

\bibitem[{Elnaggar et~al.(2021)Elnaggar, Heinzinger, Dallago, Rehawi, Wang, Jones, Gibbs, Feher, Angerer, Steinegger et~al.}]{elnaggar2021prottrans}
Ahmed Elnaggar, Michael Heinzinger, Christian Dallago, Ghalia Rehawi, Yu~Wang, Llion Jones, Tom Gibbs, Tamas Feher, Christoph Angerer, Martin Steinegger, and 1 others. 2021.
\newblock Prottrans: Toward understanding the language of life through self-supervised learning.
\newblock \emph{IEEE transactions on pattern analysis and machine intelligence}, 44(10):7112--7127.

\bibitem[{Fan et~al.(2025)Fan, Zhou, Wang, Yan, Liu, Zhao, Song, and Li}]{fan2025computational}
Wenqi Fan, Yi~Zhou, Shijie Wang, Yuyao Yan, Hui Liu, Qian Zhao, Le~Song, and Qing Li. 2025.
\newblock Computational protein science in the era of large language models (llms).
\newblock \emph{arXiv preprint arXiv:2501.10282}.

\bibitem[{Ferruz et~al.(2022)Ferruz, Schmidt, and H{\"o}cker}]{ferruz2022protgpt2}
Noelia Ferruz, Steffen Schmidt, and Birte H{\"o}cker. 2022.
\newblock Protgpt2 is a deep unsupervised language model for protein design.
\newblock \emph{Nature communications}, 13(1):4348.

\bibitem[{Gao et~al.(2024)Gao, Qiang, Tan, Jia, Ren, Lu, Liu, Ma, and Lan}]{gao2024drugclip}
Bowen Gao, Bo~Qiang, Haichuan Tan, Yinjun Jia, Minsi Ren, Minsi Lu, Jingjing Liu, Wei-Ying Ma, and Yanyan Lan. 2024.
\newblock Drugclip: Contrasive protein-molecule representation learning for virtual screening.
\newblock \emph{Advances in Neural Information Processing Systems}, 36.

\bibitem[{Greslehner(2018)}]{greslehner2018molecular}
Gregor~P Greslehner. 2018.
\newblock What do molecular biologists mean when they say'structure determines function'?

\bibitem[{Hamamsy et~al.(2022)Hamamsy, Morton, Berenberg, Carriero, Gligorijevic, Blackwell, Strauss, Leman, Cho, and Bonneau}]{hamamsy2022tm}
Tymor Hamamsy, James~T Morton, Daniel Berenberg, Nicholas Carriero, Vladimir Gligorijevic, Robert Blackwell, Charlie~EM Strauss, Julia~Koehler Leman, Kyunghyun Cho, and Richard Bonneau. 2022.
\newblock Tm-vec: template modeling vectors for fast homology detection and alignment.
\newblock \emph{bioRxiv}, pages 2022--07.

\bibitem[{Hamamsy et~al.(2023)Hamamsy, Morton, Blackwell, Berenberg, Carriero, Gligorijevic, Strauss, Leman, Cho, and Bonneau}]{hamamsy2023protein}
Tymor Hamamsy, James~T Morton, Robert Blackwell, Daniel Berenberg, Nicholas Carriero, Vladimir Gligorijevic, Charlie~EM Strauss, Julia~Koehler Leman, Kyunghyun Cho, and Richard Bonneau. 2023.
\newblock Protein remote homology detection and structural alignment using deep learning.
\newblock \emph{Nature biotechnology}, pages 1--11.

\bibitem[{Hashemifar et~al.(2018)Hashemifar, Neyshabur, Khan, and Xu}]{hashemifar2018predicting}
Somaye Hashemifar, Behnam Neyshabur, Aly~A Khan, and Jinbo Xu. 2018.
\newblock Predicting protein--protein interactions through sequence-based deep learning.
\newblock \emph{Bioinformatics}, 34(17):i802--i810.

\bibitem[{Hayes et~al.(2025)Hayes, Rao, Akin, Sofroniew, Oktay, Lin, Verkuil, Tran, Deaton, Wiggert et~al.}]{hayes2025simulating}
Thomas Hayes, Roshan Rao, Halil Akin, Nicholas~J Sofroniew, Deniz Oktay, Zeming Lin, Robert Verkuil, Vincent~Q Tran, Jonathan Deaton, Marius Wiggert, and 1 others. 2025.
\newblock Simulating 500 million years of evolution with a language model.
\newblock \emph{Science}, 387(6736):850--858.

\bibitem[{He et~al.(2016)He, Zhang, Ren, and Sun}]{he2016deep}
Kaiming He, Xiangyu Zhang, Shaoqing Ren, and Jian Sun. 2016.
\newblock Deep residual learning for image recognition.
\newblock In \emph{Proceedings of the IEEE conference on computer vision and pattern recognition}, pages 770--778.

\bibitem[{Hermosilla and Ropinski(2022)}]{hermosilla2022contrastive}
Pedro Hermosilla and Timo Ropinski. 2022.
\newblock Contrastive representation learning for 3d protein structures.
\newblock \emph{arXiv preprint arXiv:2205.15675}.

\bibitem[{Hie et~al.(2022)Hie, Candido, Lin, Kabeli, Rao, Smetanin, Sercu, and Rives}]{hie2022high}
Brian Hie, Salvatore Candido, Zeming Lin, Ori Kabeli, Roshan Rao, Nikita Smetanin, Tom Sercu, and Alexander Rives. 2022.
\newblock A high-level programming language for generative protein design.
\newblock \emph{bioRxiv}, pages 2022--12.

\bibitem[{Hochreiter and Schmidhuber(1997)}]{hochreiter1997long}
Sepp Hochreiter and J{\"u}rgen Schmidhuber. 1997.
\newblock Long short-term memory.
\newblock \emph{Neural computation}, 9(8):1735--1780.

\bibitem[{Hou et~al.(2018)Hou, Adhikari, and Cheng}]{hou2018deepsf}
Jie Hou, Badri Adhikari, and Jianlin Cheng. 2018.
\newblock Deepsf: deep convolutional neural network for mapping protein sequences to folds.
\newblock \emph{Bioinformatics}, 34(8):1295--1303.

\bibitem[{Hu et~al.(2023)Hu, Liu, Yu, and Perez-Beltrachini}]{hu2023improving}
Hanxu Hu, Yunqing Liu, Zhongyi Yu, and Laura Perez-Beltrachini. 2023.
\newblock Improving user controlled table-to-text generation robustness.
\newblock \emph{arXiv preprint arXiv:2302.09820}.

\bibitem[{Huang et~al.(2024)Huang, Li, Wu, Su, Lin, Zhang, Liu, Gao, Zheng, and Li}]{huang2024protein}
Yufei Huang, Siyuan Li, Lirong Wu, Jin Su, Haitao Lin, Odin Zhang, Zihan Liu, Zhangyang Gao, Jiangbin Zheng, and Stan~Z Li. 2024.
\newblock Protein 3d graph structure learning for robust structure-based protein property prediction.
\newblock In \emph{Proceedings of the AAAI Conference on Artificial Intelligence}, volume~38, pages 12662--12670.

\bibitem[{Jaeger et~al.(2018)Jaeger, Fulle, and Turk}]{jaeger2018mol2vec}
Sabrina Jaeger, Simone Fulle, and Samo Turk. 2018.
\newblock Mol2vec: unsupervised machine learning approach with chemical intuition.
\newblock \emph{Journal of chemical information and modeling}, 58(1):27--35.

\bibitem[{Jha et~al.(2022)Jha, Saha, and Singh}]{jha2022prediction}
Kanchan Jha, Sriparna Saha, and Hiteshi Singh. 2022.
\newblock Prediction of protein--protein interaction using graph neural networks.
\newblock \emph{Scientific Reports}, 12(1):8360.

\bibitem[{Jin et~al.(2024)Jin, Xue, Wang, Kang, Ye, Zhou, Du, and Zhang}]{jin2024prollm}
Mingyu Jin, Haochen Xue, Zhenting Wang, Boming Kang, Ruosong Ye, Kaixiong Zhou, Mengnan Du, and Yongfeng Zhang. 2024.
\newblock Prollm: protein chain-of-thoughts enhanced llm for protein-protein interaction prediction.
\newblock \emph{arXiv preprint arXiv:2405.06649}.

\bibitem[{Jumper et~al.(2021)Jumper, Evans, Pritzel, Green, Figurnov, Ronneberger, Tunyasuvunakool, Bates, {\v{Z}}{\'\i}dek, Potapenko et~al.}]{jumper2021highly}
John Jumper, Richard Evans, Alexander Pritzel, Tim Green, Michael Figurnov, Olaf Ronneberger, Kathryn Tunyasuvunakool, Russ Bates, Augustin {\v{Z}}{\'\i}dek, Anna Potapenko, and 1 others. 2021.
\newblock Highly accurate protein structure prediction with alphafold.
\newblock \emph{nature}, 596(7873):583--589.

\bibitem[{Kaminski et~al.(2023)Kaminski, Ludwiczak, Pawlicki, Alva, and Dunin-Horkawicz}]{kaminski2023plm}
Kamil Kaminski, Jan Ludwiczak, Kamil Pawlicki, Vikram Alva, and Stanislaw Dunin-Horkawicz. 2023.
\newblock plm-blast: distant homology detection based on direct comparison of sequence representations from protein language models.
\newblock \emph{Bioinformatics}, 39(10):btad579.

\bibitem[{Li et~al.(2018)Li, Gong, Yu, and Zhou}]{li2018deep}
Hang Li, Xiu-Jun Gong, Hua Yu, and Chang Zhou. 2018.
\newblock Deep neural network based predictions of protein interactions using primary sequences.
\newblock \emph{Molecules}, 23(8):1923.

\bibitem[{Li et~al.(2024{\natexlab{a}})Li, Li, Liu, Zhou, and Li}]{li2024tomg}
Jiatong Li, Junxian Li, Yunqing Liu, Dongzhan Zhou, and Qing Li. 2024{\natexlab{a}}.
\newblock Tomg-bench: Evaluating llms on text-based open molecule generation.
\newblock \emph{arXiv preprint arXiv:2412.14642}.

\bibitem[{Li et~al.(2024{\natexlab{b}})Li, Liu, Fan, Wei, Liu, Tang, and Li}]{li2024empowering}
Jiatong Li, Yunqing Liu, Wenqi Fan, Xiao-Yong Wei, Hui Liu, Jiliang Tang, and Qing Li. 2024{\natexlab{b}}.
\newblock Empowering molecule discovery for molecule-caption translation with large language models: A chatgpt perspective.
\newblock \emph{IEEE transactions on knowledge and data engineering}, 36(11):6071--6083.

\bibitem[{Li et~al.(2024{\natexlab{c}})Li, Liu, Liu, Le, Zhang, Fan, Zhou, Li, and Li}]{li2024molreflect}
Jiatong Li, Yunqing Liu, Wei Liu, Jingdi Le, Di~Zhang, Wenqi Fan, Dongzhan Zhou, Yuqiang Li, and Qing Li. 2024{\natexlab{c}}.
\newblock Molreflect: Towards in-context fine-grained alignments between molecules and texts.
\newblock \emph{arXiv preprint arXiv:2411.14721}.

\bibitem[{Li et~al.(2025)Li, Guo, Shi, Wang, and Li}]{li2025efficient}
Yiran Li, Gongyao Guo, Jieming Shi, Sibo Wang, and Qing Li. 2025.
\newblock Efficient integration of multi-view attributed graphs for clustering and embedding.
\newblock In \emph{2025 IEEE 41st International Conference on Data Engineering (ICDE)}, pages 3863--3875. IEEE.

\bibitem[{Liao et~al.(2023)Liao, Luo, Li, and Shi}]{liao2023ld2}
Ningyi Liao, Siqiang Luo, Xiang Li, and Jieming Shi. 2023.
\newblock Ld2: Scalable heterophilous graph neural network with decoupled embeddings.
\newblock \emph{Advances in neural information processing systems}, 36:10197--10209.

\bibitem[{Lieu et~al.(2020)Lieu, Nguyen, Rhyne, and Kim}]{lieu2020amino}
Elizabeth~L Lieu, Tu~Nguyen, Shawn Rhyne, and Jiyeon Kim. 2020.
\newblock Amino acids in cancer.
\newblock \emph{Experimental \& molecular medicine}, 52(1):15--30.

\bibitem[{Lin et~al.(2023)Lin, Tao, Li, and Huang}]{lin2023protein}
Peicong Lin, Huanyu Tao, Hao Li, and Sheng-You Huang. 2023.
\newblock Protein--protein contact prediction by geometric triangle-aware protein language models.
\newblock \emph{Nature Machine Intelligence}, 5(11):1275--1284.

\bibitem[{Lipman and Pearson(1985)}]{lipman1985rapid}
David~J Lipman and William~R Pearson. 1985.
\newblock Rapid and sensitive protein similarity searches.
\newblock \emph{Science}, 227(4693):1435--1441.

\bibitem[{Liu et~al.(2023)Liu, Fan, Liu, Li, Li, Liu, Tang, and Li}]{liu2023generative}
Chengyi Liu, Wenqi Fan, Yunqing Liu, Jiatong Li, Hang Li, Hui Liu, Jiliang Tang, and Qing Li. 2023.
\newblock Generative diffusion models on graphs: methods and applications.
\newblock In \emph{Proceedings of the Thirty-Second International Joint Conference on Artificial Intelligence}, pages 6702--6711.

\bibitem[{Liu et~al.(2022)Liu, Luo, Uchino, Maruhashi, and Ji}]{liu2022generating}
Meng Liu, Youzhi Luo, Kanji Uchino, Koji Maruhashi, and Shuiwang Ji. 2022.
\newblock Generating 3d molecules for target protein binding.
\newblock \emph{arXiv preprint arXiv:2204.09410}.

\bibitem[{Lopez and Mohiuddin(2024)}]{lopez2024biochemistry}
Michael~J Lopez and Shamim~S Mohiuddin. 2024.
\newblock Biochemistry, essential amino acids.
\newblock In \emph{StatPearls [Internet]}. StatPearls Publishing.

\bibitem[{Lv et~al.(2021)Lv, Hu, Bi, and Zhang}]{lv2021learning}
Guofeng Lv, Zhiqiang Hu, Yanguang Bi, and Shaoting Zhang. 2021.
\newblock Learning unknown from correlations: Graph neural network for inter-novel-protein interaction prediction.
\newblock \emph{arXiv preprint arXiv:2105.06709}.

\bibitem[{Madani et~al.(2020)Madani, McCann, Naik, Keskar, Anand, Eguchi, Huang, and Socher}]{madani2020progen}
Ali Madani, Bryan McCann, Nikhil Naik, Nitish~Shirish Keskar, Namrata Anand, Raphael~R Eguchi, Po-Ssu Huang, and Richard Socher. 2020.
\newblock Progen: Language modeling for protein generation.
\newblock \emph{arXiv preprint arXiv:2004.03497}.

\bibitem[{Mirdita et~al.(2022)Mirdita, Sch{\"u}tze, Moriwaki, Heo, Ovchinnikov, and Steinegger}]{mirdita2022colabfold}
Milot Mirdita, Konstantin Sch{\"u}tze, Yoshitaka Moriwaki, Lim Heo, Sergey Ovchinnikov, and Martin Steinegger. 2022.
\newblock Colabfold: making protein folding accessible to all.
\newblock \emph{Nature methods}, 19(6):679--682.

\bibitem[{Moal and Fern{\'a}ndez-Recio(2012)}]{moal2012skempi}
Iain~H Moal and Juan Fern{\'a}ndez-Recio. 2012.
\newblock Skempi: a structural kinetic and energetic database of mutant protein interactions and its use in empirical models.
\newblock \emph{Bioinformatics}, 28(20):2600--2607.

\bibitem[{Ofer et~al.(2021)Ofer, Brandes, and Linial}]{ofer2021language}
Dan Ofer, Nadav Brandes, and Michal Linial. 2021.
\newblock The language of proteins: Nlp, machine learning \& protein sequences.
\newblock \emph{Computational and Structural Biotechnology Journal}, 19:1750--1758.

\bibitem[{Peng et~al.(2022)Peng, Luo, Guan, Xie, Peng, and Ma}]{peng2022pocket2mol}
Xingang Peng, Shitong Luo, Jiaqi Guan, Qi~Xie, Jian Peng, and Jianzhu Ma. 2022.
\newblock Pocket2mol: Efficient molecular sampling based on 3d protein pockets.
\newblock In \emph{International Conference on Machine Learning}, pages 17644--17655. PMLR.

\bibitem[{Rao et~al.(2019)Rao, Bhattacharya, Thomas, Duan, Chen, Canny, Abbeel, and Song}]{rao2019evaluating}
Roshan Rao, Nicholas Bhattacharya, Neil Thomas, Yan Duan, Peter Chen, John Canny, Pieter Abbeel, and Yun Song. 2019.
\newblock Evaluating protein transfer learning with tape.
\newblock \emph{Advances in neural information processing systems}, 32.

\bibitem[{R{\'e}au et~al.(2023)R{\'e}au, Renaud, Xue, and Bonvin}]{reau2023deeprank}
Manon R{\'e}au, Nicolas Renaud, Li~C Xue, and Alexandre~MJJ Bonvin. 2023.
\newblock Deeprank-gnn: a graph neural network framework to learn patterns in protein--protein interfaces.
\newblock \emph{Bioinformatics}, 39(1):btac759.

\bibitem[{Renaud et~al.(2021)Renaud, Geng, Georgievska, Ambrosetti, Ridder, Marzella, R{\'e}au, Bonvin, and Xue}]{renaud2021deeprank}
Nicolas Renaud, Cunliang Geng, Sonja Georgievska, Francesco Ambrosetti, Lars Ridder, Dario~F Marzella, Manon~F R{\'e}au, Alexandre~MJJ Bonvin, and Li~C Xue. 2021.
\newblock Deeprank: a deep learning framework for data mining 3d protein-protein interfaces.
\newblock \emph{Nature communications}, 12(1):7068.

\bibitem[{Rives et~al.(2021)Rives, Meier, Sercu, Goyal, Lin, Liu, Guo, Ott, Zitnick, Ma et~al.}]{rives2021biological}
Alexander Rives, Joshua Meier, Tom Sercu, Siddharth Goyal, Zeming Lin, Jason Liu, Demi Guo, Myle Ott, C~Lawrence Zitnick, Jerry Ma, and 1 others. 2021.
\newblock Biological structure and function emerge from scaling unsupervised learning to 250 million protein sequences.
\newblock \emph{Proceedings of the National Academy of Sciences}, 118(15):e2016239118.

\bibitem[{Rocklin et~al.(2017)Rocklin, Chidyausiku, Goreshnik, Ford, Houliston, Lemak, Carter, Ravichandran, Mulligan, Chevalier et~al.}]{rocklin2017global}
Gabriel~J Rocklin, Tamuka~M Chidyausiku, Inna Goreshnik, Alex Ford, Scott Houliston, Alexander Lemak, Lauren Carter, Rashmi Ravichandran, Vikram~K Mulligan, Aaron Chevalier, and 1 others. 2017.
\newblock Global analysis of protein folding using massively parallel design, synthesis, and testing.
\newblock \emph{Science}, 357(6347):168--175.

\bibitem[{Scholkopf et~al.(1997)Scholkopf, Sung, Burges, Girosi, Niyogi, Poggio, and Vapnik}]{scholkopf1997comparing}
Bernhard Scholkopf, Kah-Kay Sung, Christopher~JC Burges, Federico Girosi, Partha Niyogi, Tomaso Poggio, and Vladimir Vapnik. 1997.
\newblock Comparing support vector machines with gaussian kernels to radial basis function classifiers.
\newblock \emph{IEEE transactions on Signal Processing}, 45(11):2758--2765.

\bibitem[{Singh et~al.(2022)Singh, Litfin, Singh, Paliwal, and Zhou}]{singh2022spot}
Jaspreet Singh, Thomas Litfin, Jaswinder Singh, Kuldip Paliwal, and Yaoqi Zhou. 2022.
\newblock Spot-contact-lm: improving single-sequence-based prediction of protein contact map using a transformer language model.
\newblock \emph{Bioinformatics}, 38(7):1888--1894.

\bibitem[{Somnath et~al.(2021)Somnath, Bunne, and Krause}]{somnath2021multi}
Vignesh~Ram Somnath, Charlotte Bunne, and Andreas Krause. 2021.
\newblock Multi-scale representation learning on proteins.
\newblock \emph{Advances in Neural Information Processing Systems}, 34:25244--25255.

\bibitem[{Su et~al.(2023)Su, Han, Zhou, Shan, Zhou, and Yuan}]{su2023saprot}
Jin Su, Chenchen Han, Yuyang Zhou, Junjie Shan, Xibin Zhou, and Fajie Yuan. 2023.
\newblock Saprot: Protein language modeling with structure-aware vocabulary. biorxiv.

\bibitem[{Suzek et~al.(2007)Suzek, Huang, McGarvey, Mazumder, and Wu}]{suzek2007uniref}
Baris~E Suzek, Hongzhan Huang, Peter McGarvey, Raja Mazumder, and Cathy~H Wu. 2007.
\newblock Uniref: comprehensive and non-redundant uniprot reference clusters.
\newblock \emph{Bioinformatics}, 23(10):1282--1288.

\bibitem[{Suzek et~al.(2015)Suzek, Wang, Huang, McGarvey, Wu, and Consortium}]{suzek2015uniref}
Baris~E Suzek, Yuqi Wang, Hongzhan Huang, Peter~B McGarvey, Cathy~H Wu, and UniProt Consortium. 2015.
\newblock Uniref clusters: a comprehensive and scalable alternative for improving sequence similarity searches.
\newblock \emph{Bioinformatics}, 31(6):926--932.

\bibitem[{Torrisi et~al.(2020)Torrisi, Pollastri, and Le}]{torrisi2020deep}
Mirko Torrisi, Gianluca Pollastri, and Quan Le. 2020.
\newblock Deep learning methods in protein structure prediction.
\newblock \emph{Computational and Structural Biotechnology Journal}, 18:1301--1310.

\bibitem[{Tsaban et~al.(2022)Tsaban, Varga, Avraham, Ben-Aharon, Khramushin, and Schueler-Furman}]{tsaban2022harnessing}
Tomer Tsaban, Julia~K Varga, Orly Avraham, Ziv Ben-Aharon, Alisa Khramushin, and Ora Schueler-Furman. 2022.
\newblock Harnessing protein folding neural networks for peptide--protein docking.
\newblock \emph{Nature communications}, 13(1):176.

\bibitem[{Unsal et~al.(2022)Unsal, Atas, Albayrak, Turhan, Acar, and Do{\u{g}}an}]{unsal2022learning}
Serbulent Unsal, Heval Atas, Muammer Albayrak, Kemal Turhan, Aybar~C Acar, and Tunca Do{\u{g}}an. 2022.
\newblock Learning functional properties of proteins with language models.
\newblock \emph{Nature Machine Intelligence}, 4(3):227--245.

\bibitem[{Vaswani et~al.(2017)Vaswani, Shazeer, Parmar, Uszkoreit, Jones, Gomez, Kaiser, and Polosukhin}]{vaswani2017attention}
Ashish Vaswani, Noam Shazeer, Niki Parmar, Jakob Uszkoreit, Llion Jones, Aidan~N Gomez, {\L}ukasz Kaiser, and Illia Polosukhin. 2017.
\newblock Attention is all you need.
\newblock \emph{Advances in neural information processing systems}, 30.

\bibitem[{Verkuil et~al.(2022)Verkuil, Kabeli, Du, Wicky, Milles, Dauparas, Baker, Ovchinnikov, Sercu, and Rives}]{verkuil2022language}
Robert Verkuil, Ori Kabeli, Yilun Du, Basile~IM Wicky, Lukas~F Milles, Justas Dauparas, David Baker, Sergey Ovchinnikov, Tom Sercu, and Alexander Rives. 2022.
\newblock Language models generalize beyond natural proteins.
\newblock \emph{BioRxiv}, pages 2022--12.

\bibitem[{Wang et~al.(2019)Wang, You, Yang, Li, Jiang, and Zhou}]{wang2019high}
Yanbin Wang, Zhu-Hong You, Shan Yang, Xiao Li, Tong-Hai Jiang, and Xi~Zhou. 2019.
\newblock A high efficient biological language model for predicting protein--protein interactions.
\newblock \emph{Cells}, 8(2):122.

\bibitem[{Wang et~al.(2022)Wang, Combs, Brand, Calvo, Xu, Price, Golovach, Salawu, Wise, Ponnapalli et~al.}]{wang2022lm}
Zichen Wang, Steven~A Combs, Ryan Brand, Miguel~Romero Calvo, Panpan Xu, George Price, Nataliya Golovach, Emmanuel~O Salawu, Colby~J Wise, Sri~Priya Ponnapalli, and 1 others. 2022.
\newblock Lm-gvp: an extensible sequence and structure informed deep learning framework for protein property prediction.
\newblock \emph{Scientific reports}, 12(1):6832.

\bibitem[{Weng et~al.(2021)Weng, Ding, Liu, Song, Chan, and Chiang}]{weng2021late}
Yue Weng, Bo~Ding, Yunqing Liu, Chunlan Song, Lo-Ying Chan, and Chien-Wei Chiang. 2021.
\newblock Late-stage photoredox c--h amidation of n-unprotected indole derivatives: Access to n-(indol-2-yl) amides.
\newblock \emph{Organic Letters}, 23(7):2710--2714.

\bibitem[{Wu et~al.(2024)Wu, Tian, Huang, Li, Lin, Chawla, and Li}]{wu2024mape}
Lirong Wu, Yijun Tian, Yufei Huang, Siyuan Li, Haitao Lin, Nitesh~V Chawla, and Stan~Z Li. 2024.
\newblock Mape-ppi: Towards effective and efficient protein-protein interaction prediction via microenvironment-aware protein embedding.
\newblock \emph{arXiv preprint arXiv:2402.14391}.

\bibitem[{Xiao et~al.(2021)Xiao, Qiu, Li, Hsieh, and Tang}]{xiao2021modeling}
Yijia Xiao, Jiezhong Qiu, Ziang Li, Chang-Yu Hsieh, and Jie Tang. 2021.
\newblock Modeling protein using large-scale pretrain language model.
\newblock \emph{arXiv preprint arXiv:2108.07435}.

\bibitem[{Xu and Zhang(2010)}]{xu2010significant}
Jinrui Xu and Yang Zhang. 2010.
\newblock How significant is a protein structure similarity with tm-score= 0.5?
\newblock \emph{Bioinformatics}, 26(7):889--895.

\bibitem[{Xu and Bonvin(2024)}]{xu2024deeprank}
Xiaotong Xu and Alexandre~MJJ Bonvin. 2024.
\newblock Deeprank-gnn-esm: a graph neural network for scoring protein--protein models using protein language model.
\newblock \emph{Bioinformatics advances}, 4(1):vbad191.

\bibitem[{Ying et~al.(2021)Ying, Cai, Luo, Zheng, Ke, He, Shen, and Liu}]{ying2021transformers}
Chengxuan Ying, Tianle Cai, Shengjie Luo, Shuxin Zheng, Guolin Ke, Di~He, Yanming Shen, and Tie-Yan Liu. 2021.
\newblock Do transformers really perform badly for graph representation?
\newblock \emph{Advances in neural information processing systems}, 34:28877--28888.

\bibitem[{Zackova~Suchanova et~al.(2023)Zackova~Suchanova, Bilcke, Romanowska, Fatlawi, Pippel, Skeffington, Schroeder, Vyverman, Vandepoele, Kr{\"o}ger et~al.}]{zackova2023diatom}
Jirina Zackova~Suchanova, Gust Bilcke, Beata Romanowska, Ali Fatlawi, Martin Pippel, Alastair Skeffington, Michael Schroeder, Wim Vyverman, Klaas Vandepoele, Nils Kr{\"o}ger, and 1 others. 2023.
\newblock Diatom adhesive trail proteins acquired by horizontal gene transfer from bacteria serve as primers for marine biofilm formation.
\newblock \emph{New Phytologist}, 240(2):770--783.

\bibitem[{Zhang et~al.()Zhang, Bi, Liang, Cheng, Hong, Deng, Zhang, Lian, and Chen}]{zhangontoprotein}
Ningyu Zhang, Zhen Bi, Xiaozhuan Liang, Siyuan Cheng, Haosen Hong, Shumin Deng, Qiang Zhang, Jiazhang Lian, and Huajun Chen.
\newblock Ontoprotein: Protein pretraining with gene ontology embedding.
\newblock In \emph{International Conference on Learning Representations}.

\bibitem[{Zhang and Skolnick(2004)}]{zhang2004scoring}
Yang Zhang and Jeffrey Skolnick. 2004.
\newblock Scoring function for automated assessment of protein structure template quality.
\newblock \emph{Proteins: Structure, Function, and Bioinformatics}, 57(4):702--710.

\bibitem[{Zhang et~al.(2022)Zhang, Xu, Jamasb, Chenthamarakshan, Lozano, Das, and Tang}]{zhang2022protein}
Zuobai Zhang, Minghao Xu, Arian Jamasb, Vijil Chenthamarakshan, Aurelie Lozano, Payel Das, and Jian Tang. 2022.
\newblock Protein representation learning by geometric structure pretraining.
\newblock \emph{arXiv preprint arXiv:2203.06125}.

\bibitem[{Zhao et~al.(2024)Zhao, Fan, Bai, Ma, Wang, Havl{\'\i}k, Cui, Balkovic, Herrero, Shi et~al.}]{zhao2024holistic}
Hao Zhao, Xiangwen Fan, Zhaohai Bai, Lin Ma, Chao Wang, Petr Havl{\'\i}k, Zhenling Cui, Juraj Balkovic, Mario Herrero, Zhou Shi, and 1 others. 2024.
\newblock Holistic food system innovation strategies can close up to 80\% of china’s domestic protein gaps while reducing global environmental impacts.
\newblock \emph{Nature Food}, pages 1--11.

\bibitem[{Zhao et~al.(2020)Zhao, Cao, and Zhang}]{zhao2020exploring}
Jingtian Zhao, Yang Cao, and Le~Zhang. 2020.
\newblock Exploring the computational methods for protein-ligand binding site prediction.
\newblock \emph{Computational and structural biotechnology journal}, 18:417--426.

\bibitem[{Zhou et~al.(2023{\natexlab{a}})Zhou, Fu, Zhang, Cheng, and Yu}]{zhou2023protein}
Hong-Yu Zhou, Yunxiang Fu, Zhicheng Zhang, Bian Cheng, and Yizhou Yu. 2023{\natexlab{a}}.
\newblock Protein representation learning via knowledge enhanced primary structure reasoning.
\newblock In \emph{The Eleventh International Conference on Learning Representations}.

\bibitem[{Zhou et~al.(2024)Zhou, Ding, Shi, Li, and Shen}]{zhou2024tined}
Ziang Zhou, Zhihao Ding, Jieming Shi, Qing Li, and Shiqi Shen. 2024.
\newblock Tined: Gnns-to-mlps by teacher injection and dirichlet energy distillation.
\newblock \emph{arXiv preprint arXiv:2412.11180}.

\bibitem[{Zhou et~al.(2023{\natexlab{b}})Zhou, Shi, Yang, Zou, and Li}]{zhou2023slotgat}
Ziang Zhou, Jieming Shi, Renchi Yang, Yuanhang Zou, and Qing Li. 2023{\natexlab{b}}.
\newblock Slotgat: slot-based message passing for heterogeneous graphs.
\newblock In \emph{International Conference on Machine Learning}, pages 42644--42657. PMLR.

\end{thebibliography}

\appendix

\clearpage
\newpage
\section{Appendix}
\label{Appendix}

\textbf{Table of Content:}
\begin{itemize}
    \item Appendix \ref{tape_results} Results on TAPE Benchmark
    \item Appendix \ref{app:ss} Secondary structure prediction
    \item Appendix \ref{app:homo} Homology Detection, Fluorescence and Stability Prediction
    \item Appendix \ref{app:function} Protein Function Prediction    
    \item Appendix \ref{app:affinity} Protein-Protein Binding Affinity Estimation
    \item Appendix \ref{ablation} Ablation Study
    \item Appendix \ref{parameter} Parameter Sensitivity Study
    \item Appendix \ref{visualization} Visualization of Protein Representations
    \item Appendix \ref{complexity} Time Complexity Analysis
    \item Appendix \ref{hyper-finetuning} Hyper-parameters for Fine-tuning
\end{itemize}

\subsection{Results on TAPE Benchmark}
\label{tape_results}

In addition to the Figure version, we also provide results on TAPE benchmark in a tabular version. As shown in Table \ref{tape}, our model GLProtein performs competitive performance on many tasks, especially on the contact prediction and stability prediction tasks. 

\subsection{Secondary structure prediction}
\label{app:ss}

% \noindent \textbf{Overview of secondary structure prediction.} 

\noindent \textbf{Overview.} Secondary structure is a fundamental aspect of computational biology, aiming to determine the local structures of protein segments. this task is a sequence-to-sequence task where each input protein is mapped to a type of local structure. We report accuracy on a per-amino acid basis on the CB513 dataset~\cite{cuff1999evaluation}.  
\noindent \textbf{Baselines.} We evaluate our model compared with ten baselines. Specifically, we employed variations of LSTM~\cite{hochreiter1997long}, ResNet~\cite{he2016deep} and Transformer~\cite{vaswani2017attention} proposed by the TAPE benchmark~\cite{rao2019evaluating}. ProtBert~\cite{elnaggar2021prottrans} is a BERT-like model pre-trained on UniRef100~\cite{suzek2007uniref,suzek2015uniref}. ESM-2~\cite{rives2021biological,verkuil2022language,hie2022high} feature a transformer architecture pre-trained on the representative sequences from UniRef50~\cite{suzek2007uniref,suzek2015uniref}. OntoProtein~\cite{zhangontoprotein} and KeAP~\cite{zhou2023protein} are the most recent knowledge-based pre-training methodologies. SaProt~\cite{su2023saprot} is the most recent structure-based protein language model. LM-GVP~\cite{wang2022lm} and GearNet~\cite{zhang2022protein} are famous geometric methods for protein representation learning.

\noindent  \textbf{Results.} For the secondary structure (SS-Q3 and SS-Q8), as shown in Figure \ref{fig:tape}, GLProtein outperforms other baselines in SS-Q8 task and shows competitive performance with ProtBERT, OntoProtein and KeAP in SS-Q3 task. Considering the approaches taken by Saprot, LM-GVP, and GearNet, which also incorporate protein structural information, the evident performance superiority of GLProtein over these methods indicates that it offers a more effective option for structure-based protein representation learning. We attribute the enhancements in performance achieved by GLProtein to its innovative integration of global and local structural information, which allows the pre-trained model to gain a deeper understanding of protein structure.

\subsection{Homology Detection, Fluorescence and Stability Prediction}
\label{app:homo}
 \noindent \textbf{Overview of homology detection.} The task of predicting remote homology in proteins can be viewed as a classification problem at the molecular level. The objective is to input a protein sequence into the homology detection model, which then identifies the correct types of protein fold. In our paper, this presents a significant challenge with 1,195 distinct protein folds to classify. We utilize data sources from ~\cite{hou2018deepsf} and present the average accuracy achieved on the fold-level heldout set. 

\noindent  \textbf{Overview of fluorescence prediction.} In the realm of protein science, fluorescence prediction is a vital task that involves estimating the fluorescence properties of proteins. This is a regression task where each input protein is mapped to a label measuring the most extreme circumstances in which the protein maintains its fold above a concentration threshold. We use the data from ~\cite{rocklin2017global} and use Spearman's rank correlation coefficient as the metric. 

\noindent  \textbf{Overview of stability prediction.} Stability prediction involves estimating the resilience of a protein's structure under various environmental conditions, a critical factor in understanding its functional efficacy and therapeutic potential. This regression task focuses on predicting the intrinsic stability of proteins, which is essential for assessing their capacity to preserve their structural integrity under severe conditions. To assess the effectiveness of our model, we measure its performance using Spearman's rank correlation coefficient across the entire test set~\cite{rocklin2017global}. 

\noindent  \textbf{Baselines.} As shown in Figure \ref{fig:tape}, we included ten protein model as baselines. 

\noindent  \textbf{Results.} As for fluorescence prediction, Figure \ref{fig:tape} shows our model has the most competitive performance compared to Transformer and KeAP. 
In the domain of stability prediction, our model again shows the highest performance with a score of 0.81. This is significantly higher compared to other models, indicating its potential utility in applications like drug design and protein engineering, where stability is paramount. 

\subsection{Protein Function Prediction}\ 
\label{app:function}

\noindent \textbf{Overview.} Protein function prediction aims to assign biological or biochemical roles to proteins, and we also regard this task as a sequence classification task. Following KeAP~\cite{zhou2023protein}, we divide protein attributes into three groups: molecular function (MF), biological process (BP) and cellular component (CC), and report the Spearman's rank correlation scores for each group. 

\noindent \textbf{Baselines.} We evaluate our model compared with five baselines, including ESM-2, ProtBERT,  OntoProtein, SaProt and KeAP. 

\begin{table}[htp]

\centering
% \vskip -0.150in
\scalebox{0.8}
{
\begin{tabular}{l c c c c}
\toprule
Methods & MF & BP & CC & Avg\\
% \toprule
\midrule
ESM-2 & 0.31 & \textbf{0.42}& 0.28& 0.34\\
ProtBert & \textbf{0.41} & 0.35& 0.36& 0.37\\
OntoProtien & \textbf{0.41} & 0.36& 0.36& 0.38\\
SaProt & \underline{0.40} & \underline{0.40}& \underline{0.39}& \textbf{0.40}\\
KeAP & \underline{0.40} & \underline{0.40}& \textbf{0.40}& \textbf{0.40}\\
\rowcolor{gray!20}
\textbf{GLProtein} & \textbf{0.41} & \underline{0.40}& \underline{0.39} & \textbf{0.40}\\
\bottomrule
\end{tabular}
}
\caption{Comparisons on semantic similarity inference. The best results are bolded, and the second-best results are underlined.
}
\label{semantic}
\end{table}

\vskip -0.150in
\noindent  \textbf{Results.} Table \ref{semantic} assesses the performance of various computational models in prediction protein functions in three categories: MF, BP and CC. Additionally, an average score (Avg) is calculated for each method to provide a holistic view of performance across all categories. These models all show a balanced performance in three groups. It is worth noting that our model does not use any protein attribute-related knowledge and is comparable to OntoProtein and KeAP, which do. It also demonstrates the superiority of our approach.

\subsection{Protein-Protein Binding Affinity Estimation}
\label{app:affinity}

\noindent \textbf{Overview.} In this task, we focus on assessing how well protein representations can predict changes in binding affinity caused by protein mutations. This regression task involves associating each protein pair with a numerical value. Following the methodology described in ~\cite{unsal2022learning}, we employ Bayesian ridge regression on the outcomes of element-wise multiplication of representations derived from pre-trained protein models. This approach is designed to enhance the accuracy of binding affinity predictions. We used the SKEMPI dataset from ~\cite{moal2012skempi} and reported the mean square error of 10-fold cross-validation.

\noindent \textbf{Baselines.} 
% \wq{we can detail more here since we don't have any space limitations.}
We evaluate our model compared with six baselines. Specifically, we employed PIPR~\cite{chen2019multifaceted}, ProtBert~\cite{elnaggar2021prottrans}, ESM-2~\cite{rives2021biological,verkuil2022language,hie2022high}, SaProt~\cite{su2023saprot}, OntoProtein~\cite{zhangontoprotein} and KeAP~\cite{zhou2023protein}. PIPR is a siamese-residual-RCNN-based model for multifaceted protein–protein interaction prediction. ProtBert is a BERT-like model pre-trained on UniRef100~\cite{suzek2007uniref,suzek2015uniref}. ESM-2 feature a transformer architecture pre-trained on the representative sequences from UniRef50~\cite{suzek2007uniref,suzek2015uniref}. SaProt is the most recent structure-based protein language model. OntoProtein and KeAP are the most recent knowledge-based pre-training methodologies.

\begin{table}[htp]
 
\centering
\begin{tabular}{l c}
\toprule
Methods & Affinity($\downarrow$) \\
% \toprule
\midrule
PIPR & 0.63\\
ProtBert & 0.58\\
ESM-2 & \textbf{0.48}\\
SaProt & 0.58\\
OntoProtien & 0.59\\
KeAP & \underline{0.52}\\
\rowcolor{gray!20}
\textbf{GLProtein} & \underline{0.52}\\
\bottomrule
\end{tabular}
\caption{Comparisons on protein-protein binding affinity prediction, with the best result bolded and the second best underlined. The notion $\downarrow$ signifies a preference for lower values, reflecting a superior predictive performance in this context. 
}
\label{affinity}
\end{table}

\begin{table*}[h]

\centering
\scalebox{0.8}{
\begin{tabular}{l c c c c}
\toprule
 & Parameters & Resouces & Pre-training & Inference\\
 &  &  &  & (40 examples)\\
% \toprule
\midrule
ProtBert & 400M & A single TPU Pod V3-512 & 400k steps & 2.02s \\
OntoProtein & 400M & 4 NVIDIA 48G A6000 GPUs & 3 Days (continue pertaining on ProtBert) & 1.91s\\
KeAP & 400M & 4 NVIDIA 48G A6000 GPUs & 3 Days (continue pertaining on ProtBert) & 1.94s\\
SaProt & 650M & 64 NVIDIA 80G A100 GPUs & 3 Months& 3.02s\\
ESM-2 & 650M & - & - & 2.45s\\
GLProtein & 400M & 4 NVIDIA 48G A6000 GPUs & 3 Days (continue pertaining on ProtBert) & 1.93s\\
\bottomrule
\end{tabular}}
\caption{Comparison of the number of parameters, resources, pre-training time, and inference time for GLProtein and baselines.
}
\label{time}
\end{table*}

\noindent  \textbf{Results.} Table \ref{affinity} compares several methods of predicting the binding affinity of protein interactions, where a lower score indicates superior performance. GLProtein outperforms PIPR, ProtBert, SaProt and OntoProtein. It also shows the competitive performance of KeAP and ESM-2. 

\subsection{Ablation Study}
\label{ablation}

\begin{figure}[h]
    \centering
    \vspace{-5mm}
    \includegraphics[width=1\linewidth]{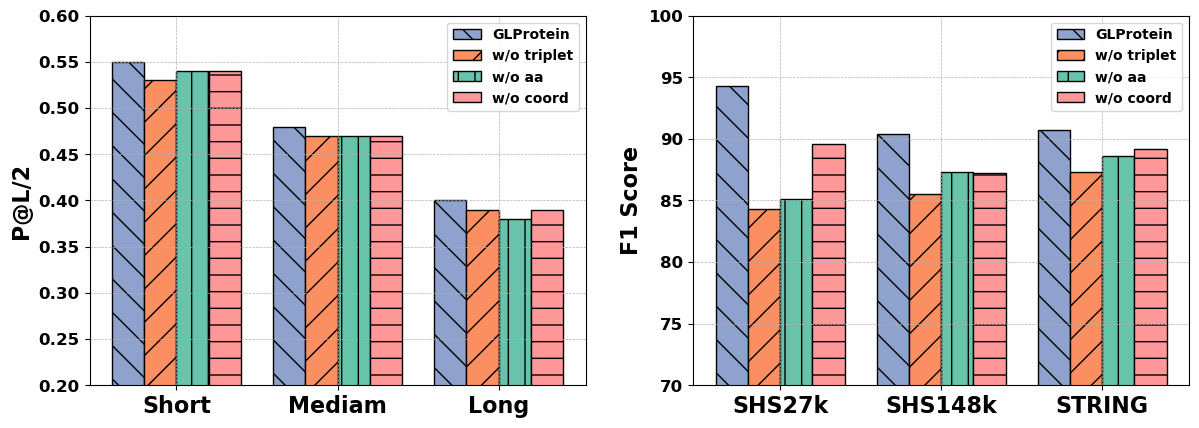}
    \vspace{-5mm}
    \caption{\textbf{Left}: Ablations of three proposed approaches. Long-range P@L/2 results are reported for contact prediction. Right: Ablations of three proposed approaches. F1 scores are reported for protein-protein interaction tasks.}
    \label{fig:ablation_study}
\end{figure}

We investigated the effects of employing diverse protein structure information fusion strategies. First of all, the exclusion of the global structure information modelling component (representation as "\textit{w/o triplet}" in Figure \ref{fig:ablation_study}) resulted in varying degrees of performance deterioration across contact prediction and protein-protein interaction prediction tasks. This observation suggests that our global structure similarities through protein triplet contrastive learning stand out as a more efficacious choice. Subsequently, upon removing the proposed substructure-based molecular encoding from the local protein structure information component (denoted as "\textit{w/o aa}" in Figure \ref{fig:ablation_study}), we noted a decline in performance by approximately 2.5\% and 8\% for contact prediction and protein-protein interaction tasks, respectively. This underscores the essential role of substructure-based molecular encoding within our proposed methodologies. Finally when the protein 3D distance encoding was omitted from the local structure information modelling component (indicated as "\textit{w/o coord}" in Figure \ref{fig:ablation_study}), a similar trend of performance degradation was observed, further emphasizing the indispensability of this strategy within our architectural framework.

\subsection{Parameter Sensitivity Study}
\label{parameter}
In this section, we explore the impact of the parameters in the model on the final performance of our protein model. We experimented with the number of protein samples in the protein local structure information modelling component and the coefficient of contrastive learning loss for the protein triplet, respectively.

\begin{figure*}
    \centering
    \includegraphics[width=1\linewidth]{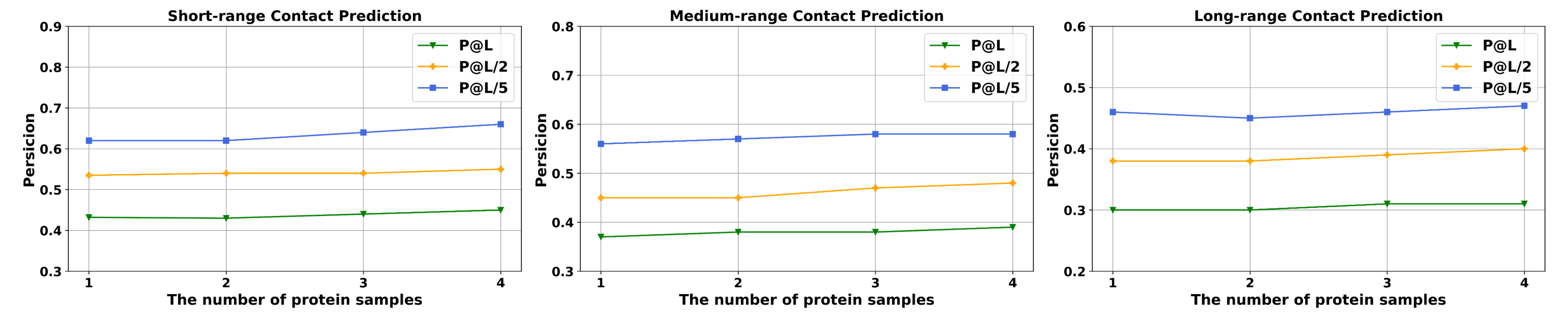}
    \caption{Parameter sensitivity study on the number of protein samples in the local structure information component.}
    \label{fig:ablation_sample}
\end{figure*}

As shown in Figure \ref{fig:ablation_sample}, we test the number of protein samples from 1 to 4 on the contact prediction task. We observe that as the number of protein samples increases, the performance of our model improves to varying degrees in short-, medium- and long-range contact prediction. This also shows that our proposed protein triplet approach indeed enables the protein language model to capture the structural similarity features among proteins. Due to computational and memory cost considerations, we ended up constructing 4 protein positive samples and 4 protein negative samples for each protein. 

\begin{figure*}
    \centering
    \includegraphics[width=1\linewidth]{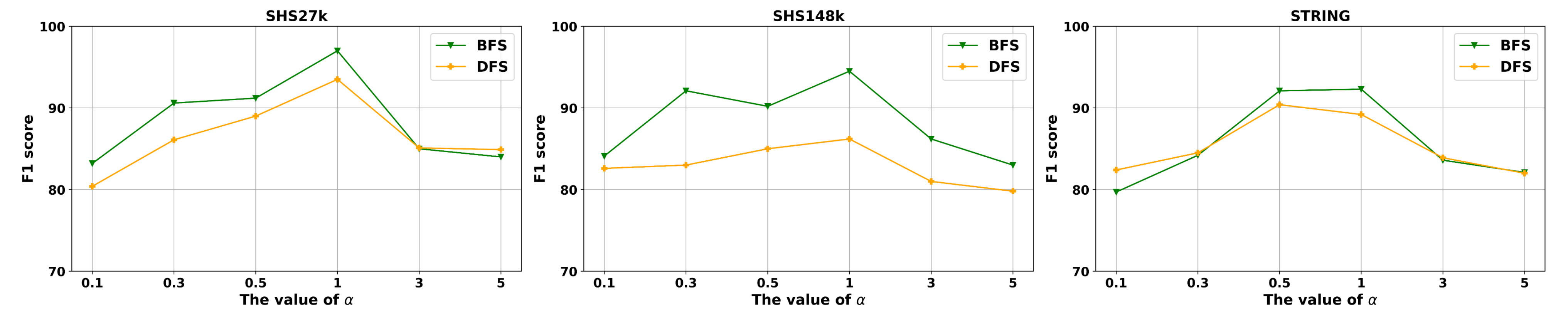}
    \caption{Parameter sensitivity study on the value of the coefficient $\alpha$ of contrastive learning loss for the protein triplet in the local structure information component.}
    \label{fig:alpha}
\end{figure*}

As shown in Figure \ref{fig:alpha}, we test the value of the coefficient $\alpha$ of contrastive learning loss for the protein triplet. We divided the experiment into 6 groups and set the values of $\alpha$ to 0.1, 0.3, 0.5, 1, 3, and 5. Then, we evaluated them using the protein-protein interaction prediction task on SHS27k, SHS148k, and STRING datasets, respectively. We observe that the model achieves the best performance when the value of $\alpha$ is set to 1. Thus, we choose $\alpha=1$ in this paper as our model's setting.

\begin{table*}[htbp]

\centering
\vspace{-0.1in}
\scalebox{0.8}{
\begin{tabular}{l c c c c c c}
\toprule
\multirow{2}{*}{}     & \multicolumn{3}{c}{Structure} & \multicolumn{1}{c}{Evolutionary} & \multicolumn{2}{c}{Engineering} \\
                      & SS-Q3 & SS-Q8 & Contact & Homology & Fluorescence & Stability     \\
\midrule
SaProt       & 0.51   & 0.45   & 0.37   & 0.12    & 0.25   & 0.46   \\
LSTM        & 0.75   & 0.59   & 0.26   & 0.26    & 0.67   & 0.69   \\
Transformer & 0.73   & 0.59   & 0.25   & 0.21    & \textbf{0.68}   & 0.73   \\
ResNet     & 0.75   & 0.58   & 0.25   & 0.17    & 0.21   & 0.73   \\
ESM-2       & 0.70 & 0.54 & 0.43 & 0.10 & 0.30 & 0.65 \\
LM-GVP       & 0.69 & 0.50 & 0.43 & 0.20 & 0.64  & 0.69 \\
GearNet       & 0.71 & 0.55 & 0.44 & 0.25 & \underline{0.67} & \underline{0.78} \\
ProtBert       & \textbf{0.82} & \underline{0.67} & 0.35 & \textbf{0.29} & 0.61 & 0.73  \\
OntoProtein       & \textbf{0.82} & \underline{0.67} & 0.40 & 0.24 & 0.65 & 0.74  \\
KeAP       & \textbf{0.82} & \underline{0.67} & \underline{0.45} & \textbf{0.29} & \underline{0.67} & 0.75  \\
\rowcolor{gray!20}
\textbf{GLProtein}       & \textbf{0.82} & \textbf{0.68} & \textbf{0.48} & \underline{0.28} & \underline{0.67} & \textbf{0.81} \\
\bottomrule
\end{tabular}}
\caption{
Results on TAPE Benchmark. SS is a secondary structure task that is evaluated in CB315. In contact prediction, we test medium- and long-range using P@L/2 metrics. In protein engineering tasks, we test fluorescence and stability prediction using Spearman's $\rho$ metric.}
\label{tape}
\end{table*}

\begin{table*}[htp]

\centering
% \vskip -0.150in
\scalebox{1}
{
\begin{tabular}{l c c c c c c c}
\toprule
Task & Epoch & Batch Size & Warmup Ratio &  Learning Rate & Freeze Bert & Optimizer\\
% \toprule
\midrule
Contact & 5 & 8& 0.08 & 3e-5 & False & AdamW \\
Homology	&10	&32	&0.08&	4e-5&	False&	AdamW \\
Stability&	5&	64&	0.08&	1e-5	&False&	AdamW \\
SS-Q3	&5	&32	&0.08	&3e-5	&False	&AdamW \\
SS-Q8	&5	&32	&0.08&	3e-5	&False	&AdamW \\
Fluorescence	&15	&64	&0.10	&1e-3	&True	&Adam \\
\bottomrule
\end{tabular}
}
\caption{Hyper-parameters for fine-tuning.
}
\label{tab:fine-hyper}
\end{table*}

\begin{figure*}[h]
	
	\begin{minipage}{0.45\linewidth}
		\vspace{3pt}

		\centerline{\includegraphics[width=\textwidth]{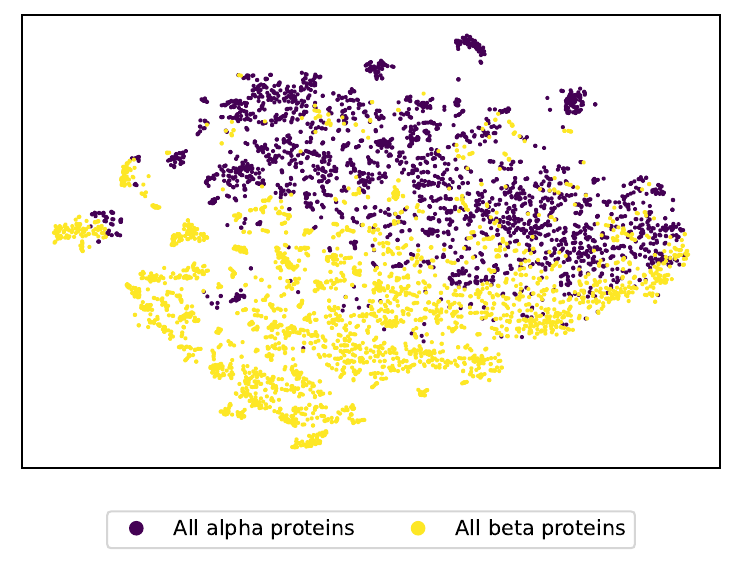}}

		\centerline{Our Proposed GLProtein}
	\end{minipage}
	\begin{minipage}{0.45\linewidth}
		\vspace{3pt}
		\centerline{\includegraphics[width=\textwidth]{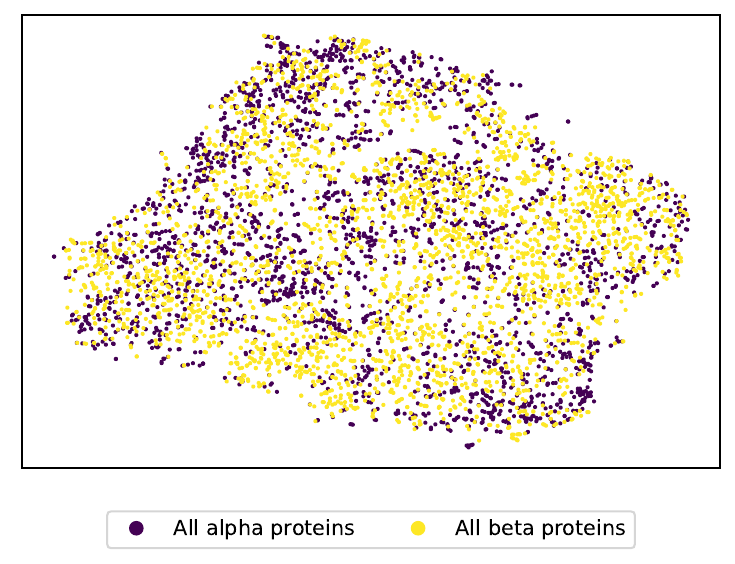}}
		\centerline{ESM-2}
	\end{minipage}

	\begin{minipage}{0.45\linewidth}
		\vspace{3pt}
		\centerline{\includegraphics[width=\textwidth]{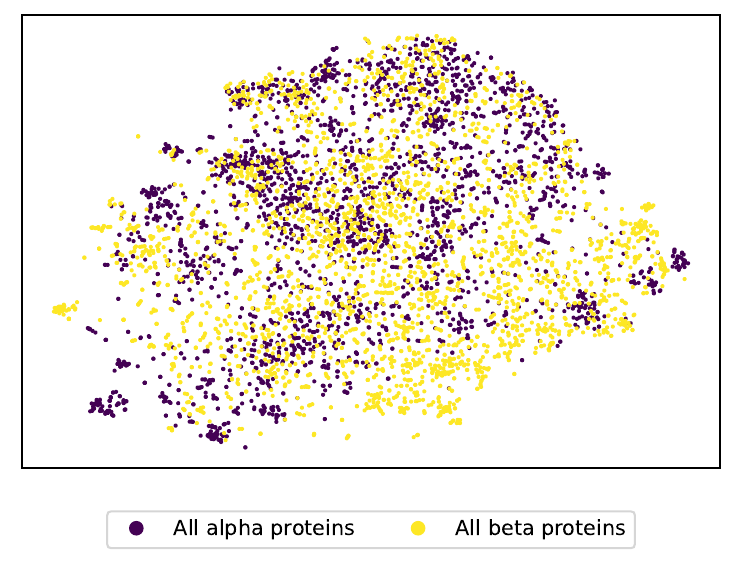}}
	 
		\centerline{KeAP}
	\end{minipage}
        \begin{minipage}{0.45\linewidth}
		\vspace{3pt}
		\centerline{\includegraphics[width=\textwidth]{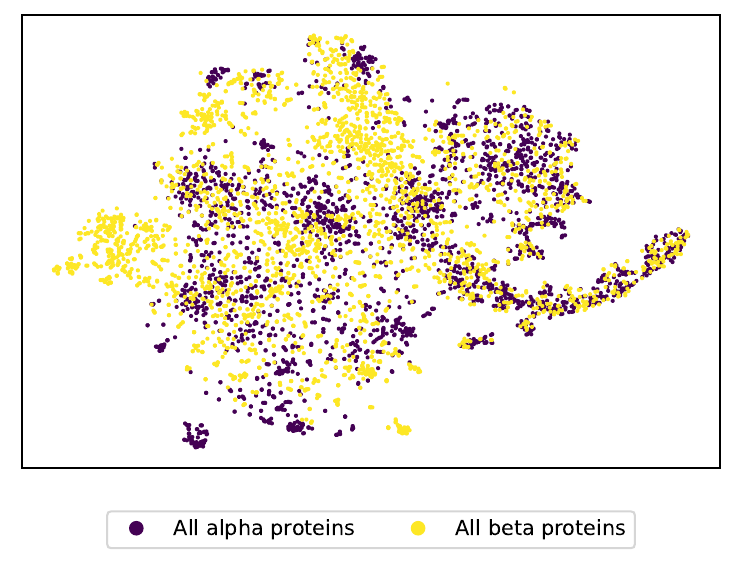}}
	 
		\centerline{ProtBert}
	\end{minipage}

        \caption{Embedding visualizations of GLProtein, ESM-2, KeAP and ProtBert on SCOPe database.}
	\label{embeddings}
\end{figure*}

\subsection{Visualization of Protein Representations}
\label{visualization}
To facilitate a more intuitive comparison, we utilize t-SNE to visualize the protein representations produced by GLProtein, ESM2, KeAP, and ProtBert. The visualization results, based on the non-redundant subset ($PIDE \le 40\%$) of the SCOPe database~\cite{chandonia2019scope}, are illustrated in Figure \ref{embeddings}. As depicted in this figure, the representations for alpha and beta proteins generated by GLProtein are distinctly separated, whereas those produced by ESM-2, KeAP, and ProtBert are more closely intertwined.

\subsection{Time Complexity Analysis}
\label{complexity}
We provide a more specific complexity analysis as follows: protein encoder operates at approximately $O(L^2d)$, where $L$ is the length of protein sequence and $d$ is the embedding dimension. Triplet protein sampling operates at approximately $O(L^3)$, reducing the complexity to $O(L^2)$ by TM-Vec. Triplet loss operates at approximately $O(3Ld)\rightarrow O(Ld) $.  Protein 3D distance encoding operated at approximately $O(KL^2d)$, where $K$ is the number of Gaussian Basis kernels. Substructure-based molecular encoding operates at approximately $O(Ld)$. Protein decoder operates at approximately  $O(L^2d)$. Total computation cost operated at $O_{total} = O_{encoder}+O_{global}+O_{local}+O_{decoder} = O(L^2d) + O(L^2)+O(Ld)+O(KL^2d)+O(L^2d) = O((K+1)L^2d) $.

\subsection{Hyper-parameters for Fine-tuning}
\label{hyper-finetuning}

The hyper-parameters for fine-tuning are provided in the Table \ref{tab:fine-hyper}. Specifically, we follow the hyper-parameter settings in GNN-PPI~\cite{lv2021learning} for PPI prediction. For protein binding affinity prediction and semantic similarity inference, we follow the fine-tuning configuration in PROBE~\cite{unsal2022learning}.

\end{document}